\documentclass[11pt]{article}

\usepackage[final]{acl}
\usepackage{times}
\usepackage{latexsym}

\usepackage[T1]{fontenc}

\usepackage[utf8]{inputenc}

\usepackage{microtype}

\usepackage{inconsolata}

\usepackage{graphicx}

\usepackage{algorithm}
\usepackage{enumitem}
\usepackage{newfloat}
\usepackage{listings}
\usepackage[most]{tcolorbox}
\usepackage{xcolor}  
\usepackage{booktabs}
\usepackage{multirow}
\usepackage{amsmath, amssymb}
\usepackage{booktabs}
\usepackage[table]{xcolor}
\usepackage{tabularx}
\usepackage{longtable}
\usepackage{array} 
\usepackage{pifont}
\usepackage{algorithm}
\usepackage{algpseudocode} 
\definecolor{yzbtodo}{RGB}{30,144,255}

\usepackage{wrapfig}
\usepackage{graphicx} 
\usepackage{xspace}
\usepackage{enumitem}
\usepackage{booktabs}         
\usepackage{multirow}         
\usepackage{colortbl}         
\usepackage[table]{xcolor}    
\usepackage{graphicx}         
\usepackage{algorithm} 
\usepackage{algpseudocode} 
\usepackage{amsmath, amsfonts, amssymb} 
\usepackage{amsthm}

\usepackage{booktabs} 
\usepackage{adjustbox} 
\usepackage{times}
\usepackage{latexsym}
\usepackage{colortbl}
\usepackage{xcolor}
\usepackage{threeparttable}
\usepackage{subcaption}  
\usepackage{threeparttable}
\usepackage{hyperref}
\theoremstyle{remark}

\definecolor{citecolor}{RGB}{2, 56, 189}
\definecolor{myred}{RGB}{207,62,62}
\definecolor{mygreen}{RGB}{112,173,71}
\definecolor{mynewred}{RGB}{178, 56, 40}
\definecolor{rebuttal_blue}{HTML}{1685a9}
\definecolor{rebuttal_purple}{HTML}{9932cd}
\definecolor{rebuttal_red}{HTML}{ef7a82}
\definecolor{avgblue}{RGB}{210,230,250}

\newcommand{\hlfirst}[1]{\colorbox[HTML]{CFE2FF}{#1}}  
\newcommand{\hlsecond}[1]{\colorbox[HTML]{F0F6FF}{#1}}

\usepackage[dvipsnames]{xcolor}

\renewcommand{\hlfirst}[1]{\textbf{#1}}
\renewcommand{\hlsecond}[1]{\underline{#1}}

\newcommand{\mname}{{\fontfamily{lmtt}\selectfont \textbf{ToolAtlas}}\xspace} 

\usepackage{booktabs}
\usepackage[table]{xcolor}
\usepackage{colortbl}
\usepackage{multirow}

\usepackage{cleveref}

\usepackage{xcolor}

\usepackage{pifont}

\title{\mname{}: Learning Once, Reusing Everywhere with Tool-Side Memory}

\author{
  \makebox[\textwidth][c]{%
    Yue Fang\textsuperscript{$\spadesuit$}\thanks{These authors contributed equally.}, 
    Zhibang Yang\textsuperscript{$\spadesuit$}\footnotemark[1], 
    Fangkai Yang\textsuperscript{$\clubsuit$}\thanks{Corresponding author.}, 
    Xiaoting Qin\textsuperscript{$\clubsuit$}\footnotemark[2], 
  } \\[4pt]
  \makebox[\textwidth][c]{%
  Liqun Li\textsuperscript{$\clubsuit$}\footnotemark[2],
    Qingwei Lin\textsuperscript{$\clubsuit$}, 
    Saravan Rajmohan\textsuperscript{$\clubsuit$}, 
    Dongmei Zhang\textsuperscript{$\clubsuit$}} \\[6pt]
  \makebox[\textwidth][c]{\textsuperscript{$\spadesuit$}School of Computer Science, Peking University, Beijing, China} \\
  \makebox[\textwidth][c]{\textsuperscript{$\clubsuit$}Microsoft} \\
}

\begin{document}
\maketitle
\begin{abstract}
Large language model (LLM) agents increasingly rely on external tools served by shared providers and accessed by heterogeneous downstream agents. Existing approaches improve tool use on the agent side through parameter updates, prompt refinement, or agent-side memory, making tool knowledge difficult to share and limited to behaviors observed in past tasks.
We argue that reusable tool knowledge should instead be maintained by the tool provider. We introduce \mname{}, a graph-based framework that builds a persistent provider-side tool memory of tool capabilities, failure boundaries, and cross-tool compositions through execution-verified probing. At inference time, agents query the tool memory via adaptive graph traversal.
Across two MCP-based benchmarks spanning eight services, \mname{} outperforms existing tool-side optimization and agent-side memory baselines by up to 21.61\% in pass@1 and 18.61\% in pass@4. The same tool memory also transfers across environment instances and agent frameworks without retraining or task-time exploration, yielding up to 24.16\%/16.22\% and 17.49\%/14.27\% relative gains in pass@1/pass@4, respectively. Ablation studies show that these gains arise from combining tool-centered memory organization with capability-guided execution probing. These results establish provider-side tool memory as an effective and reusable paradigm for tool servers. Our code is in: {\url{https://github.com/PuppyKnightUniversity/ToolAtlas}.
}
\end{abstract}

\section{Introduction}\label{sec:introduction}
Tool use has become a defining capability of modern large language model~(LLM) agents, enabling them to invoke external services and data sources beyond their parametric knowledge. Recent work shows that such tool use now spans large, realistic ecosystems, from comprehensive tool-augmented benchmarks to thousands of real-world APIs~\cite{DBLP:conf/emnlp/LiZ000YLHL23,DBLP:conf/nips/PatilZ0G24, qin2024toolllm}. Under the Model Context Protocol~(MCP), this ecosystem is increasingly deployed in a server-centric form: tool providers expose capabilities through MCP servers, while heterogeneous agents consume them across products and environments~\cite{mcp2026intro, openai2026mcpconnectors}. Official provider-hosted servers from GitHub and Stripe further reflect this
trend~\cite{github2026mcpserver, stripe2026mcp}. In this setting, the same tool server may be called by many downstream agents with different prompts, action interfaces, and runtime environments.


\begin{figure}[t]
    \centering
    \includegraphics[scale=0.15]{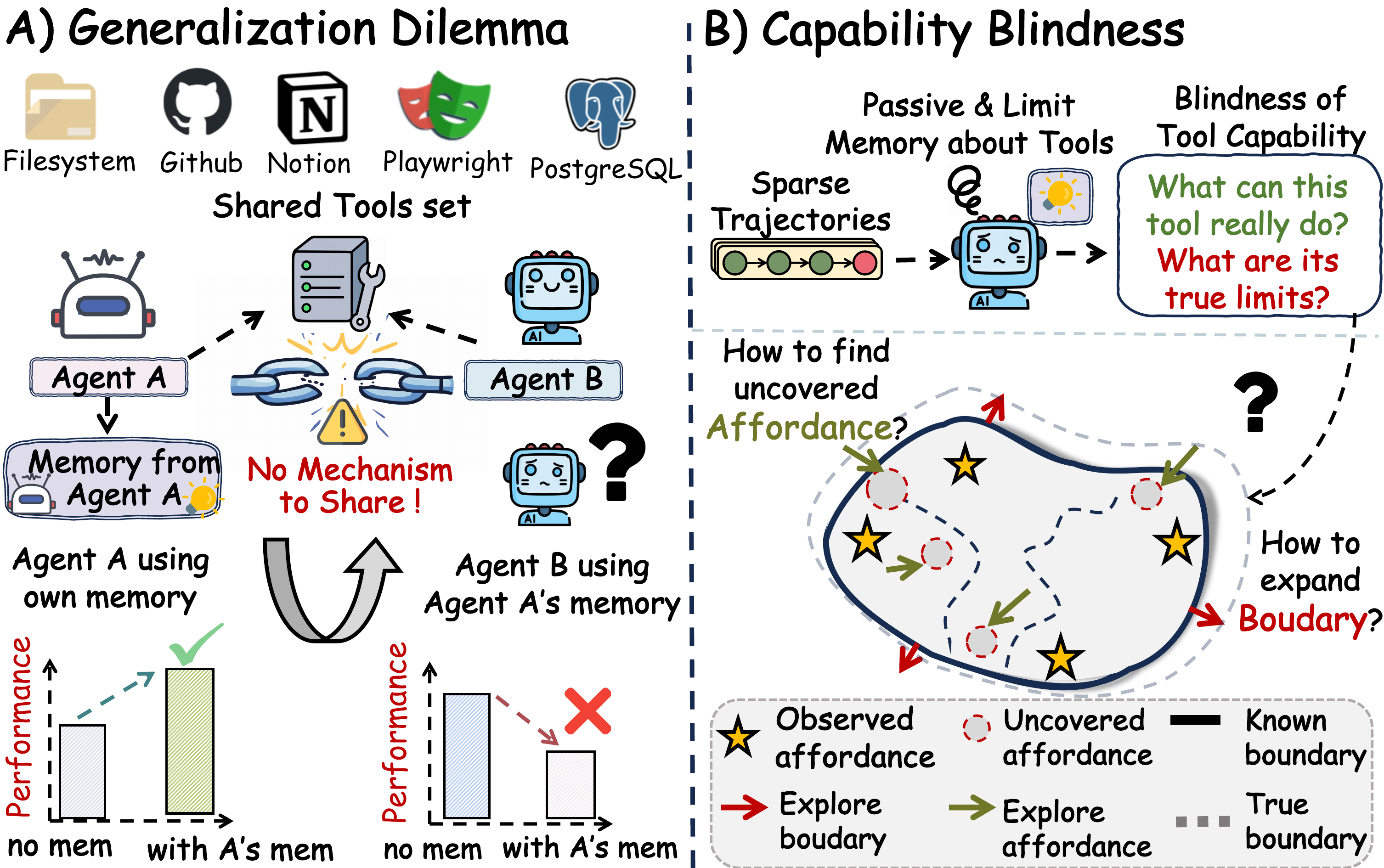}
    \caption{Two structural limitations of agent-side tool memory in shared tool server settings.} 
    \vspace{-1em}
    \label{fig:introduction}
\end{figure}


Despite this shared-server setting, most prior work still improves tool use on the agent side. Agent-side methods internalize tool-use behavior into model parameters~\cite{schick2023toolformer,qin2024toolllm,
liu2025toolace}, create reusable skills or tools~\cite{wang2023voyager,cai2024large,qian2023creator,yuan2024craft},
or accumulate experience in agent-side memory~\cite{zhao2024expel,
ouyang2025reasoningbank, xu2026mem, zhang2025agentic, xiao2025toolmem}. Existing tool-side efforts refine tool descriptions, documentation, retrieval, or catalog organization~\cite{yuan2025easytool,
fang2025play2prompt, wu2025joint, qu2025exploration, liu2025toolscope}. However, existing tool-side optimization mainly improves how tools are presented or selected, while agent-side memory leaves knowledge inside the agent that produced it, rather than making it reusable across agents that use same tool.

This ownership mismatch matters in shared-server deployments. When many agents access one MCP server, each agent must rediscover behaviors and failure modes already observed by others. From the provider's perspective, this repeated exploration is wasteful: a single offline effort could be amortized across downstream agents if the resulting knowledge traveled with the tool. Moreover, passively logging agent trajectories reveals only the subset of tool behavior exercised by past tasks. These observations lead to two challenges:

\begin{itemize}
[leftmargin=*,itemsep=0pt,parsep=0pt,topsep=0pt,partopsep=0pt]

    \item \textit{\textbf{Generalization Dilemma}.} Agent-side memory offers no mechanism for sharing what one agent learns with another. Each entry is written in the producing agent's own workflow, action format, reasoning style, and environment assumptions, so the same lesson must be relearned by every agent that calls the tool, even when the underlying tool behavior is unchanged (Figure~\ref{fig:introduction}(A)). Tool knowledge is therefore
    entangled with agent identity and difficult to reuse across agents.

    \item \textit{\textbf{Capability Blindness}.} Agent-side memory records only the tool behaviors exercised by past tasks. It follows the task distribution rather than the tool's capability, so untriggered boundary conditions and unused tool compositions are absent by construction (Figure~\ref{fig:introduction}(B)). The result is a sparse task footprint rather than a deliberate account of what the tool can do, where it fails, and how it composes with others.
    
\end{itemize}

These challenges suggest a two-part answer. Tool memory should be provider-side rather than agent-side, and it should be built through deliberate exploration of each tool's capability frontier rather than passive logging of tasks. To this end, we propose \mname{}, which maintains a provider-side tool memory graph for each tool server. The graph records what a tool can do, where it fails, and how it composes with peers, written once for any agent to consult. Because the graph is maintained with the tool, its construction cost can be amortized across many downstream agents. The graph has three interlinked layers. A Tool-Trace Graph distills past task executions into agent-neutral tool-rationale traces. A Tool-Capability Graph attaches affordance and boundary entries to each tool and connects tools through co-usage edges. A Tool-Strategy Graph extracts recurring orchestration principles from many tasks. To extend the graph along each tool's frontier rather than along incidental task coverage, \mname{} replaces passive trajectory collection with Frontier Exploration, an iterative loop in which an LLM probes the graph's current gaps along two complementary directions, outward toward untested boundary conditions and inward toward unverified affordances and tool compositions, executes each probe against the live tool, and folds the verified outcome back into the graph to sharpen next round. At inference, a lightweight navigator traverses the graph adaptively, expanding relevant regions step by step instead of retrieving a static top-$k$ slice.
Our contributions are threefold:
\begin{itemize}[leftmargin=*,itemsep=0pt,parsep=0pt,topsep=0pt,partopsep=0pt]
    \item We argue that tool memory should live with the tool rather than inside individual agents, so reusable tool knowledge is no longer tied to one agent's workflow or action format.
    
    \item We introduce \mname{}, a provider-side tool-memory framework that organizes knowledge as a tool memory graph, expands it through capability exploration, and retrieves it through adaptive graph traversal, making tool experience reusable across agents and less tied to past task coverage.
    
    \item Experiments on MCPMark and MCP-Universe across eight MCP services show that \mname{} outperforms tool-side optimization and agent-side memory baselines, while transferring across agents and task categories without re-training or task-time exploration.
\end{itemize}

\section{Related Work}
\label{sec:related}


\noindent\textbf{Agent-Side Tool-Use Improvement.}
Most prior work improves tool use on the agent side. One line
internalizes tool-use behavior into model parameters through
pre-training, supervised fine-tuning, or reinforcement learning~\cite{schick2023toolformer, qin2024toolllm, liu2025toolace}. Another line lets agents expand their own action space by creating reusable tools,
programs, or skills, as in Voyager~\cite{wang2023voyager} and
tool-creation frameworks~\cite{cai2024large,qian2023creator,
yuan2024craft}. Closest to our setting are agent-side memory methods. ExpeL~\cite{zhao2024expel} distills past trials into
natural-language insights for later tasks. A-MEM~\cite{xu2026mem}
organizes memories as dynamically linked notes. ACE~\cite{zhang2025agentic}
treats context as an evolving playbook; and ToolMem learns
tool-capability memories for multimodal agents~\cite{xiao2025toolmem}. These methods improve agent performance, but the acquired knowledge remains tied to a particular model, skill library, or agent trajectory history, making it difficult to share across heterogeneous agents that access the same tool server.

\noindent\textbf{Tool-Side Efforts and Interface Optimization.}
A complementary line improves how tools are exposed, selected, or
organized for LLM agents. Tool documentation can serve as an effective
zero-shot interface~\cite{hsieh2023tool}, motivating methods that
optimize tool descriptions and usage examples, including
EasyTool~\cite{yuan2025easytool}, DRAFT~\cite{qu2025exploration},
Play2Prompt~\cite{fang2025play2prompt}, and
ToolOptimal~\cite{wu2025joint}. Other work handles large tool catalogs
by retrieving, filtering, or structuring candidates, as in
Re-Invoke~\cite{chen2024reinvoke}, ToolRet~\cite{shi2025toolret},
AnyTool~\cite{du2024anytool}, ToolScope~\cite{liu2025toolscope}, and
ToolExpNet~\cite{zhang2025toolexpnet}. These efforts improve tool
interfaces or selection, but do not maintain provider-side memory of
execution-verified capabilities, failure boundaries, and cross-tool
compositions. \mname{} fills this gap with reusable tool-side memory
built through frontier-guided execution probing and shared across
heterogeneous agents without retraining or per-agent memory accumulation.

\begin{figure*}[t]
  \centering
\includegraphics[width=1.0\textwidth]{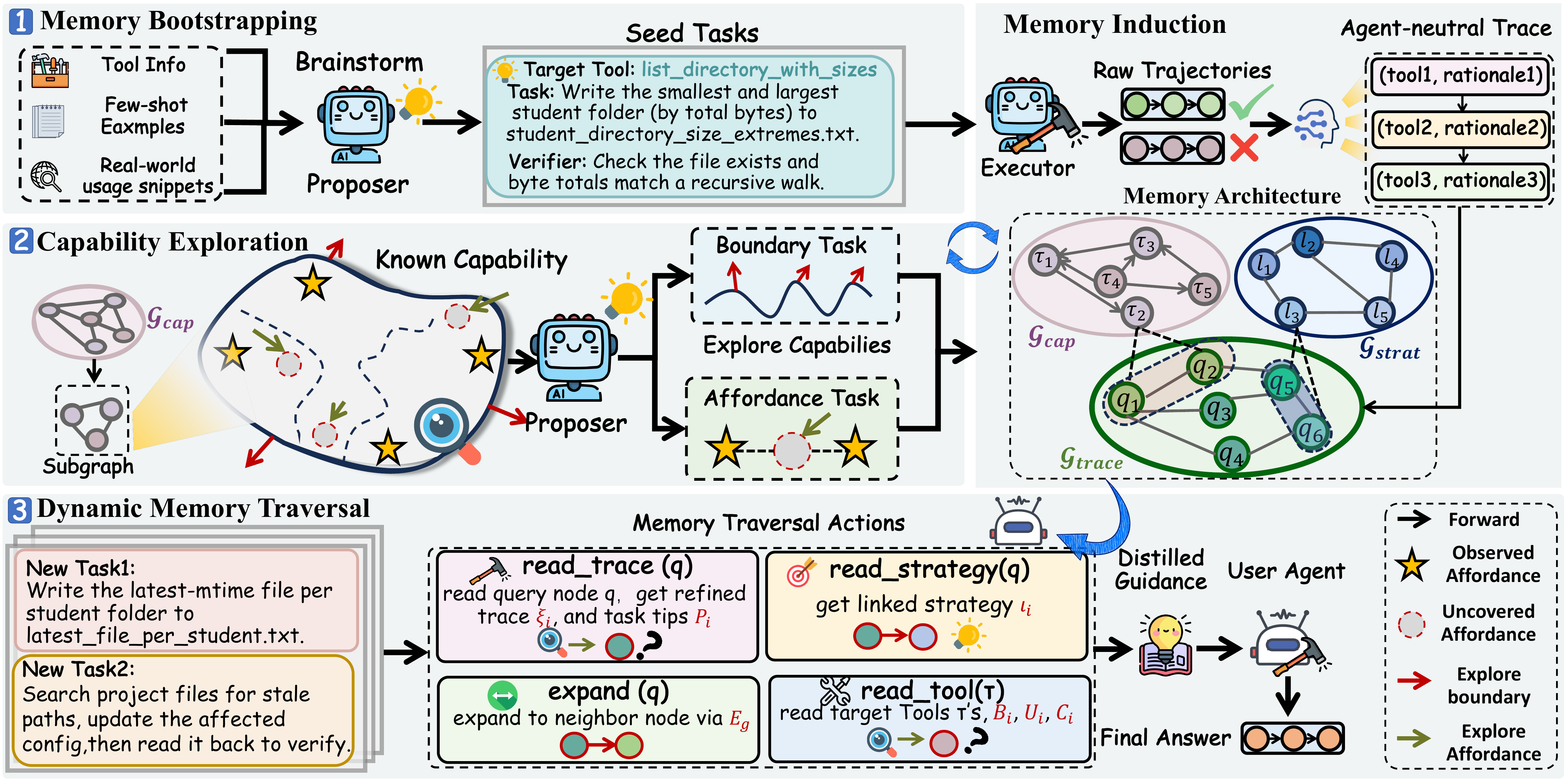}
  \caption{Overview of \mname. Stage 1 bootstraps a tool-side memory graph from seed tasks and verified rollouts, where Memory Induction converts raw trajectories into tool memory graph. Stage 2 uses the current graph to propose capability explorations that test tool boundaries and affordances, then writes verified outcomes back to the graph. Stage 3 traverses the graph at inference time and summarizes the retrieved traces, tool entries, and strategies into task-conditioned guidance for a user agent.}
  \label{fig:modelStructure}
\end{figure*}

\section{Methodology}
\label{sec:method}

\mname{} constructs a tool-side memory graph through three stages, as shown in \Cref{fig:modelStructure}. First, \textit{Memory Bootstrapping} uses tool specifications and web-retrieved usage snippets to generate seed tasks for every tool, executes the tasks with verifiers, and induces an initial graph from the resulting rollouts. Second, \textit{Capability Exploration} uses the current graph to address Capability Blindness (Section~\ref{sec:introduction}) in two directions. It pushes outward to test tool capability boundaries and fills inward to uncover under-recorded affordances, meaning valid ways to use a tool that the graph has not yet captured, together with tool compositions. The generated probes are executed and their verified outcomes are written back to the graph. Third, at inference time, \textit{Dynamic Memory Traversal} navigates the tool memory graph to gather related traces, tool entries, and strategies, then compresses the gathered information into task-conditioned guidance for the user agent. Appendix~\ref{app:case-seed-round} shows a sample case study.

\subsection{Memory Bootstrapping}
\label{sec:memory-bootstrap}

Memory Bootstrapping gives a bare MCP service an initial tool-side memory. Let $\mathcal{T}=\{t_1,\ldots,t_M\}$ denote the tools exposed by the service, and let $\mathcal{M}_{\mathcal{T}}$ denote the tool memory graph built for them. For each tool $t_i\in\mathcal{T}$, \mname{} synthesizes $k$ seed task-verifier pairs, defined as $\mathcal{D}^{(0)}_i=\{(Q_{ij},v_{ij})\}_{j=1}^{k}$, where $Q_{ij}$ is the $j$-th seed task for $t_i$ and $v_{ij}$ is its executable verifier. The full seed set is $\mathcal{D}^{(0)}=\bigcup_{i=1}^{M}\mathcal{D}^{(0)}_i$. The generated tasks are intended to cover representative uses of each tool. Since tool specifications are often too short to support diverse task synthesis, the task proposer also receives real-world usage snippets retrieved through web search. Each seed task $Q_{ij}$ is executed for $N$ independent rollouts, giving a rollout set $\mathcal{R}^{(0)}_{ij}=\{(\xi_{ij}^{(n)},y_{ij}^{(n)})\}_{n=1}^{N}$, where $\xi_{ij}^{(n)}$ is the raw trajectory and $y_{ij}^{(n)}\in\{\textsc{Resolved},\textsc{Failed}\}$ is the verifier-labeled outcome. These verified rollouts are then passed to Memory Induction.

\noindent\textbf{Memory Induction.} Memory Induction creates the tool memory graph from executed rollouts. During Memory Bootstrapping, it starts from an empty graph and incrementally inserts entries induced from each seed task and its rollout set. The motivation is that raw rollouts are still agent-specific because they contain the action syntax, retry pattern, and incidental reasoning style of the agent that produced them. Even rollouts for the same task may reveal different successful plans, repeated mistakes, or failure modes. Memory Induction therefore converts these rollouts into graph entries that keep reusable tool-use knowledge while discarding agent-specific form.

\noindent\textit{Tool-Trace Graph.} 
Memory Induction condenses multiple rollouts of the same task into an \emph{agent-neutral trace} $\tilde{\xi}_{ij}$, a sequence of \textit{(tool, rationale)} pairs that removes agent-specific action syntax while preserving the tool-use intent. Together with the task and task-level tips $\mathcal{P}_{ij}$ distilled from successful and failed rollouts, this forms a trace node $q_{ij}=(Q_{ij},\tilde{\xi}_{ij},\mathcal{P}_{ij})$.
The trace nodes form the Tool-Trace Graph $\mathcal{G}_{\mathsf{trace}}=(\mathcal{V}_{\mathsf{trace}},\mathcal{E}_{\mathsf{trace}})$, where $\mathcal{V}_{\mathsf{trace}}$ is the set of trace nodes and semantic edges in $\mathcal{E}_{\mathsf{trace}}$ connect related trace nodes by embedding similarity over task summaries. For example, noisy rollouts for \texttt{list\_directory\_with\_sizes} may contain dozens of calls and repeated replanning loops. Memory Induction compresses them into a short trace that keeps only the transferable tool-rationale sequence, such as discovering the workspace, inspecting candidate folders, checking sizes, verifying totals, and saving the final report.

\noindent\textit{Tool-Capability Graph.} Agent-neutral traces make rollouts reusable, but they are still organized by task. To support tool-side capability discovery, \mname{} builds a Tool-Capability Graph $\mathcal{G}_{\mathsf{cap}}=(\mathcal{V}_{\mathsf{tool}},\mathcal{E}_{\mathsf{tool}})$ whose nodes are tools and whose inter-tool edges summarize recurrent co-usage patterns observed in agent-neutral traces. A tool node is connected to every trace node whose agent-neutral trace uses that tool, and we denote these trace-tool links by $\mathcal{E}_{\mathsf{trace}\leftrightarrow\mathsf{tool}}$. These links let the tool node gather what related trace nodes reveal about the tool.
Each tool node $t_i$ carries three entry sets $(\mathcal{U}_i,\mathcal{B}_i,\mathcal{C}_i)$, where affordance entries $\mathcal{U}_i$ record valid uses that $t_i$ reliably supports, boundary entries $\mathcal{B}_i$ mark regimes where $t_i$ becomes unreliable or infeasible, and co-usage entries $\mathcal{C}_i$ document recurrent compositions with peer tools.
This graph is the operational state used later for capability exploration. It tells the proposer what is already known about a tool and which affordances, boundaries, or compositions remain under-covered.

\noindent\textit{Tool-Strategy Graph.} Some lessons are planning patterns rather than single-tool properties. Memory Induction captures them by grouping semantically related trace nodes or trace nodes with shared tool-use patterns, then distilling recurring task tips into strategy nodes. These nodes form the Tool-Strategy Graph $\mathcal{G}_{\mathsf{strat}}=(\mathcal{V}_{\mathsf{strat}},\mathcal{E}_{\mathsf{strat}})$. We denote bidirectional trace-strategy links by $\mathcal{E}_{\mathsf{trace}\leftrightarrow\mathsf{strat}}$. For example, file-system traces that first check allowed directories and locate the writable workspace can induce a strategy to discover the workspace and target scope before acting. This graph lets a user agent reuse high-level plans on similar tasks.

After Memory Induction, the complete tool memory graph consists of the three graphs,
$\mathcal{M}_{\mathcal{T}}=(\mathcal{G}_{\mathsf{trace}},\mathcal{G}_{\mathsf{cap}},\mathcal{G}_{\mathsf{strat}})$
with cross-graph links $\mathcal{E}_{\mathsf{trace}\leftrightarrow\mathsf{tool}}$ between trace nodes and supported tool nodes, and $\mathcal{E}_{\mathsf{trace}\leftrightarrow\mathsf{strat}}$ between trace nodes and supported strategy nodes.

\subsection{Capability Exploration}
\label{sec:capability-explore}

Memory Bootstrapping gives the MCP service an initial tool-side memory, but seed tasks still cover only a small region of the tool capability. Capability Exploration expands $\mathcal{M}_{\mathcal{T}}$ by using the Tool-Capability graph as feedback.
The graph records what is already known about each tool, and the task proposer uses that state to propose tasks that test what remains uncertain.
At round $r$, \mname{} samples a local subgraph $\mathcal{S}^{(r)}$ of $K$ related tools from the current Tool-Capability Graph $\mathcal{G}^{(r-1)}_{\mathsf{cap}}$. The sampler starts from a target tool and follows co-usage edges, favoring plausible compositions while preserving coverage of under-explored tools. 

For each target tool $t_i\in\mathcal{S}^{(r)}$, \mname{} builds a proposer context $\mathcal{K}^{(r)}_i$ from the target tool node, its entries $(\mathcal{U}_i,\mathcal{B}_i,\mathcal{C}_i)$, neighboring tools, and linked traces. Conditioned on this context, an LLM proposer generates probe tasks $\mathcal{D}^{(r)}_{i,d}$ along two directions, $d\in\{\mathsf{boundary},\mathsf{affordance}\}$. The outward direction tests capability boundaries where a tool becomes unreliable or infeasible, such as asking \texttt{list\_directory\_with\_sizes} to rank deeply nested folders by recursive byte total when its node records shallow per-entry sizes. The inward direction uncovers under-recorded affordances and compositions, such as testing whether \texttt{list\_directory\_with\_sizes} with \texttt{get\_file\_info} can find the largest file under each student directory. Each proposed task is paired with a verifier and executed by the agent. The resulting probe rollouts $\mathcal{R}^{(r)}$ are passed through Memory Induction to update $\mathcal{M}^{(r-1)}_{\mathcal{T}}$ into $\mathcal{M}^{(r)}_{\mathcal{T}}$, adding trace nodes, tool capability entries, and strategy nodes.

\subsection{Dynamic Memory Traversal}
\label{sec:traversal}

At inference time, \mname{} guides a user agent on a new task by walking the tool memory graph. Fixed top-$k$ retrieval is insufficient because evidence may span trace, tool, and strategy graph layers. Dynamic Memory Traversal performs a graph walk over $\mathcal{M}_{\mathcal{T}}$.
Given a new task $Q$, traversal retrieves the $k_r$ nearest Tool-Trace nodes as the initial context $C_0$. At step $s$, a lightweight navigator selects graph navigation operation $a_s$ updates the context as $C_{s+1}=C_s\cup\textsc{Navigate}(a_s;\mathcal{M}_{\mathcal{T}})$. 
The navigator uses four operations: (1) \textsc{ReadTrace}$(q)$ reads the task, agent-neutral trace, and tips of trace node $q$; (2) \textsc{Expand}$(q)$ follows trace-graph edges to similar trace nodes; (3) \textsc{ReadTool}$(t)$ reads affordance, boundary, and co-usage entries from tool node $t$; and (4) \textsc{ReadStrategy}$(q)$ reads strategy nodes supported by trace node $q$.
Traversal stops when the navigator judges the context sufficient or the read budget is exhausted. The final context is compressed into task-conditioned guidance $g(Q)$ and injected into the user agent's memory slot before rollout.


\begin{table*}[!ht]
\centering
\fontsize{7.5pt}{8pt}\selectfont
\setlength{\tabcolsep}{2.4pt}
\renewcommand{\arraystretch}{1.25}
\begin{tabular}{l|cc|cc|cc|cc|cc|cc}
\toprule
\multirow{2}{*}{\textbf{Method}}
& \multicolumn{2}{c|}{\textbf{Filesystem}}
& \multicolumn{2}{c|}{\textbf{GitHub}}
& \multicolumn{2}{c|}{\textbf{Notion}}
& \multicolumn{2}{c|}{\textbf{Playwright}}
& \multicolumn{2}{c|}{\textbf{PostgreSQL}}
& \multicolumn{2}{c}{\textbf{Overall}} \\
& pass@1$\uparrow$ & pass@4$\uparrow$
& pass@1$\uparrow$ & pass@4$\uparrow$
& pass@1$\uparrow$ & pass@4$\uparrow$
& pass@1$\uparrow$ & pass@4$\uparrow$
& pass@1$\uparrow$ & pass@4$\uparrow$
& pass@1$\uparrow$ & pass@4$\uparrow$ \\
\midrule
\rowcolor{gray!10}
\multicolumn{13}{c}{\textit{GPT-5.4}} \\
\midrule
Vanilla
& 48.68 & 68.42 & 46.67 & 53.33 & 44.44 & 55.56 & 46.15 & 61.54 & 59.62 & 69.23
& 49.11 & 61.62 \\
EasyTool
& 52.63 & 78.95 & 46.67 & \hlsecond{66.67} & 45.83 & \hlsecond{66.67} & 50.00 & 61.54 & 59.62 & \hlsecond{76.92}
& 50.95 & 70.15 \\
Play2Prompt
& 50.00 & 78.95 & 51.67 & \hlsecond{66.67} & 48.61 & 61.11 & 53.85 & \hlsecond{69.23} & 61.54 & \hlsecond{76.92}
& 53.13 & 70.58 \\
ToolOptimal
& 53.95 & \hlsecond{84.21} & 48.33 & 60.00 & 48.61 & \hlsecond{66.67} & 50.00 & 61.54 & 53.85 & 69.23
& 50.95 & 68.33 \\
Expel
& 48.68 & 78.95
& 36.67 & 53.33
& 44.44 & \hlsecond{66.67}
& \hlsecond{55.77} & \hlsecond{69.23}
& 55.77 & \hlsecond{76.92}
& 48.27 & 69.02 \\
A-Mem
& 57.89 & 78.95 & 46.67 & 60.00 & 43.06 & 55.56 & 42.31 & 53.85 & 59.62 & \hlfirst{84.61}
& 49.91 & 66.59 \\
ACE
& \hlsecond{59.21} & \hlsecond{84.21}
& \hlsecond{53.33} & \hlsecond{66.67}
& \hlsecond{50.00} & \hlsecond{66.67}
& 53.85 & 61.54
& \hlsecond{65.38} & \hlfirst{84.61}
& \hlsecond{56.35} & \hlsecond{72.74} \\
\rowcolor[HTML]{F0F6FF}
\mname{}
& \hlfirst{69.73} & \hlfirst{94.73} & \hlfirst{56.67} & \hlfirst{73.33} & \hlfirst{51.39} & \hlfirst{72.22} & \hlfirst{59.62} & \hlfirst{76.92} & \hlfirst{67.30} & \hlfirst{84.61}
& \hlfirst{60.94} & \hlfirst{80.36} \\
\textit{Impro(\%)}
& \textit{+17.77} & \textit{+12.49} & \textit{+6.26} & \textit{+9.99} & \textit{+2.78} & \textit{+8.32} & \textit{+6.90} & \textit{+11.11} & \textit{+2.94} & \textit{+0.00} & \textit{+8.15} & \textit{+10.48} \\
\midrule
\rowcolor{gray!10}
\multicolumn{13}{c}{\textit{GPT-5.4-mini}} \\
\midrule
Vanilla
& 15.79 & 26.32 & 15.00 & 33.33 & 26.39 & 44.44 & 15.38 & 38.46 & 46.15 & 61.54
& 23.74 & 40.82 \\
EasyTool
& 22.37 & \hlsecond{52.63} & 26.67 & \hlsecond{46.67} & 23.61 & 38.89 & 21.15 & \hlsecond{61.54} & 50.00 & 61.54
& 28.76 & 52.25 \\
Play2Prompt
& 19.74 & 47.37 & \hlsecond{28.33} & \hlsecond{46.67} & 26.39 & \hlfirst{55.56} & 17.31 & 53.85 & 50.00 & \hlsecond{69.23}
& 28.35 & \hlsecond{54.54} \\
ToolOptimal
& \hlsecond{23.68} & \hlsecond{52.63} & \hlsecond{28.33} & 40.00 & 22.22 & \hlsecond{50.00} & 15.38 & 38.46 & \hlsecond{53.85} & \hlsecond{69.23}
& 28.69 & 50.06 \\
Expel
& 17.11 & 42.11
& 26.67 & \hlsecond{46.67}
& 26.39 & 44.44
& 26.92 & 46.15
& 48.08 & 61.54
& 29.03 & 48.18 \\
A-Mem
& 21.05 & \hlsecond{52.63} & \hlsecond{28.33} & 40.00 & 23.61 & 44.44 & 23.08 & 46.15 & \hlsecond{53.85} & 61.54
& 29.98 & 48.95 \\
ACE
& \hlsecond{23.68} & \hlsecond{52.63}
& \hlsecond{28.33} & 40.00
& \hlsecond{31.94} & 44.44
& \hlsecond{32.69} & 53.85
& \hlsecond{53.85} & \hlsecond{69.23}
& \hlsecond{34.10} & 52.03 \\
\rowcolor[HTML]{F0F6FF}
\mname{}
& \hlfirst{28.94} & \hlfirst{68.42}
& \hlfirst{31.67} & \hlfirst{53.33}
& \hlfirst{34.72} & \hlfirst{55.56}
& \hlfirst{36.53} & \hlfirst{69.23}
& \hlfirst{55.77} & \hlfirst{76.92}
& \hlfirst{37.53} & \hlfirst{64.69} \\
\textit{Impro(\%)}
& \textit{+22.21} & \textit{+30.00} & \textit{+11.79} & \textit{+14.27} & \textit{+8.70} & \textit{+0.00} & \textit{+11.75} & \textit{+12.50} & \textit{+3.57} & \textit{+11.11} & \textit{+10.06} & \textit{+18.61} \\
\bottomrule
\end{tabular}
\vspace{-0.2cm}
\captionsetup{font=footnotesize}
\caption{Performance comparison on \textsc{MCPMark} under the same-environment split, using \textit{SwitchAct} with GPT-5.4 and GPT-5.4-mini as backbones. Results with Grok-4 are reported in \Cref{tab:rq1_mcpmark_grok}. The \textbf{best} and \underline{second best} results are marked in bold and with underline, respectively. \textit{Impro(\%)} is the relative gain of \mname{} over the strongest baseline.}
\label{tab:rq1_mcpmark_gpt54_gpt54mini}
\end{table*}
\begin{table*}[!ht]
\centering
\fontsize{7.5pt}{8pt}\selectfont
\setlength{\tabcolsep}{2.4pt}
\renewcommand{\arraystretch}{1.25}
\begin{tabular}{l|cc|cc|cc|cc|cc|cc}
\toprule
\multirow{2}{*}{\textbf{Method}}
& \multicolumn{2}{c|}{\textbf{Filesystem}}
& \multicolumn{2}{c|}{\textbf{GitHub}}
& \multicolumn{2}{c|}{\textbf{Notion}}
& \multicolumn{2}{c|}{\textbf{Playwright}}
& \multicolumn{2}{c|}{\textbf{PostgreSQL}}
& \multicolumn{2}{c}{\textbf{Overall}} \\
& pass@1$\uparrow$ & pass@4$\uparrow$
& pass@1$\uparrow$ & pass@4$\uparrow$
& pass@1$\uparrow$ & pass@4$\uparrow$
& pass@1$\uparrow$ & pass@4$\uparrow$
& pass@1$\uparrow$ & pass@4$\uparrow$
& pass@1$\uparrow$ & pass@4$\uparrow$ \\
\midrule
Vanilla
& 48.53 & 58.82
& 44.44 & \hlsecond{66.67}
& \hlsecond{36.76} & \hlsecond{58.82}
& 42.86 & 57.14
& 55.77 & 61.54
& 45.67 & 60.60 \\
EasyTool
& 51.47 & \hlsecond{70.59}
& 44.44 & 55.56
& 20.59 & 41.18
& \hlsecond{53.57} & \hlsecond{64.29}
& \hlsecond{57.69} & \hlsecond{69.23}
& 45.55 & 60.17 \\
Play2Prompt
& \hlsecond{52.94} & \hlfirst{76.47}
& \hlsecond{47.22} & \hlsecond{66.67}
& 25.00 & 47.06
& 50.00 & \hlsecond{64.29}
& 55.77 & 61.54
& \hlsecond{46.19} & \hlsecond{63.21} \\
ToolOptimal
& 41.18 & 64.71
& 30.56 & 55.56
& 27.94 & 52.94
& 44.64 & \hlfirst{71.43}
& 55.77 & \hlsecond{69.23}
& 40.02 & 62.77 \\
Expel
& 45.59 & \hlsecond{70.59}
& 30.56 & 44.44
& 29.41 & 47.06
& 39.29 & 57.14
& 48.08 & 61.54
& 38.59 & 56.15 \\
A-Mem
& 44.23 & 53.85
& 36.11 & 44.44
& 32.35 & 52.94
& 35.71 & \hlfirst{71.43}
& 46.15 & 61.54
& 38.91 & 56.84 \\
ACE
& 44.12 & 58.82
& 38.89 & 55.56
& 30.88 & 41.18
& 35.71 & 42.86
& 51.92 & \hlsecond{69.23}
& 40.30 & 53.53 \\
\rowcolor[HTML]{F0F6FF}
\mname{}
& \hlfirst{61.76} & \hlfirst{76.47}
& \hlfirst{58.33} & \hlfirst{77.78}
& \hlfirst{44.12} & \hlfirst{64.71}
& \hlfirst{57.14} & \hlfirst{71.43}
& \hlfirst{65.38} & \hlfirst{76.92}
& \hlfirst{57.35} & \hlfirst{73.46} \\
\textit{Impro(\%)}
& \textit{+16.66} & \textit{+0.00}
& \textit{+23.53} & \textit{+16.66}
& \textit{+20.02} & \textit{+10.01}
& \textit{+6.66} & \textit{+0.00}
& \textit{+13.33} & \textit{+11.11}
& \textit{+24.16} & \textit{+16.22} \\
\bottomrule
\end{tabular}
\vspace{-0.2cm}
\captionsetup{font=footnotesize}
\caption{Performance comparison on \textsc{MCPMark} under the cross-environment split, using \textit{SwitchAct} with GPT-5.4 as the backbone. The \textbf{best} and \underline{second best} results are marked in bold and with underline, respectively. \textit{Impro(\%)} is the relative gain of \mname{} over the strongest baseline.}
\vspace{-3mm}
\label{tab:rq2_cross_environment}
\end{table*}
\begin{table*}[!ht]
\centering
\fontsize{7.5pt}{8pt}\selectfont
\setlength{\tabcolsep}{2.4pt}
\renewcommand{\arraystretch}{1.25}
\begin{tabular}{l|cc|cc|cc|cc|cc|cc}
\toprule
\multirow{2}{*}{\textbf{Method}}
& \multicolumn{2}{c|}{\textbf{Filesystem}}
& \multicolumn{2}{c|}{\textbf{GitHub}}
& \multicolumn{2}{c|}{\textbf{Notion}}
& \multicolumn{2}{c|}{\textbf{Playwright}}
& \multicolumn{2}{c|}{\textbf{PostgreSQL}}
& \multicolumn{2}{c}{\textbf{Overall}} \\
& pass@1$\uparrow$ & pass@4$\uparrow$
& pass@1$\uparrow$ & pass@4$\uparrow$
& pass@1$\uparrow$ & pass@4$\uparrow$
& pass@1$\uparrow$ & pass@4$\uparrow$
& pass@1$\uparrow$ & pass@4$\uparrow$
& pass@1$\uparrow$ & pass@4$\uparrow$ \\
\midrule
\rowcolor{gray!10}
\multicolumn{13}{c}{\textit{Test on ReAct}} \\
\midrule
Vanilla
& 51.32 & 63.16
& \hlsecond{48.33} & 53.33
& \hlsecond{43.06} & \hlsecond{61.11}
& 32.69 & 53.85
& 48.08 & \hlsecond{69.23}
& 44.70 & 60.14 \\
EasyTool
& \hlsecond{60.53} & 73.68
& 43.33 & 46.67
& 41.67 & \hlsecond{61.11}
& 34.62 & 53.85
& 42.31 & 61.54
& 44.49 & 59.37 \\
Play2Prompt
& 52.63 & 73.68
& 41.67 & 46.67
& 41.67 & \hlfirst{66.67}
& \hlsecond{46.15} & \hlsecond{61.54}
& \hlsecond{55.77} & 61.54
& \hlsecond{47.58} & 62.02 \\
ToolOptimal
& 57.89 & \hlsecond{78.95}
& 36.67 & 40.00
& \hlsecond{43.06} & \hlfirst{66.67}
& 36.54 & 53.85
& 51.92 & 61.54
& 45.22 & 60.20 \\
Expel
& 51.32 & 73.68
& 46.67 & 53.33
& 40.28 & \hlsecond{61.11}
& 30.77 & 46.15
& 46.15 & \hlsecond{69.23}
& 43.04 & 60.70 \\
A-Mem
& 50.00 & \hlsecond{78.95}
& 45.00 & \hlsecond{60.00}
& 25.00 & \hlsecond{61.11}
& 25.00 & \hlsecond{61.54}
& 34.62 & \hlsecond{69.23}
& 35.92 & \hlsecond{66.17} \\
ACE
& 55.26 & 63.16
& 45.00 & 46.67
& 38.89 & 55.56
& 34.62 & 53.85
& 50.00 & 61.54
& 44.75 & 56.16 \\
\rowcolor[HTML]{F0F6FF}
\mname{}
& \hlfirst{64.47} & \hlfirst{84.21}
& \hlfirst{50.00} & \hlfirst{73.33}
& \hlfirst{48.61} & \hlfirst{66.67}
& \hlfirst{51.92} & \hlfirst{69.23}
& \hlfirst{61.54} & \hlfirst{84.62}
& \hlfirst{55.31} & \hlfirst{75.61} \\
\textit{Impro(\%)}
& \textit{+6.51} & \textit{+6.66}
& \textit{+3.46} & \textit{+22.22}
& \textit{+12.89} & \textit{+0.00}
& \textit{+12.50} & \textit{+12.50}
& \textit{+10.35} & \textit{+22.23}
& \textit{+16.25} & \textit{+14.27} \\
\midrule
\rowcolor{gray!10}
\multicolumn{13}{c}{\textit{Test on CodeAct}} \\
\midrule
Vanilla
& 53.95 & 68.42
& 21.67 & 40.00
& 41.67 & 61.11
& 42.31 & \hlsecond{53.85}
& 57.69 & \hlsecond{76.92}
& 43.46 & 60.06 \\
EasyTool
& \hlsecond{65.79} & \hlsecond{84.21}
& 26.67 & 33.33
& 43.06 & \hlsecond{66.67}
& 44.23 & \hlsecond{53.85}
& \hlsecond{61.54} & 69.23
& \hlsecond{48.26} & 61.46 \\
Play2Prompt
& 53.95 & 78.95
& \hlsecond{33.33} & \hlsecond{46.67}
& \hlsecond{51.39} & \hlsecond{66.67}
& 42.31 & \hlsecond{53.85}
& 57.69 & \hlsecond{76.92}
& 47.73 & \hlsecond{64.61} \\
ToolOptimal
& 64.47 & \hlsecond{84.21}
& 28.33 & 33.33
& 45.83 & 61.11
& 40.38 & \hlsecond{53.85}
& 53.85 & 61.54
& 46.57 & 58.81 \\
Expel
& 56.58 & 68.42
& 20.00 & 33.33
& 40.28 & 55.56
& 40.38 & 46.15
& 57.69 & 69.23
& 42.99 & 54.54 \\
A-Mem
& 56.58 & \hlsecond{84.21}
& 21.67 & 26.67
& 43.06 & 55.56
& \hlsecond{46.15} & \hlfirst{61.54}
& 30.77 & 61.54
& 39.65 & 57.90 \\
ACE
& 52.63 & 63.16
& 18.33 & 40.00
& 41.67 & 61.11
& 40.38 & \hlsecond{53.85}
& 55.77 & 61.54
& 41.76 & 55.93 \\
\rowcolor[HTML]{F0F6FF}
\mname{}
& \hlfirst{78.95} & \hlfirst{89.47}
& \hlfirst{38.33} & \hlfirst{53.33}
& \hlfirst{52.78} & \hlfirst{72.22}
& \hlfirst{48.08} & \hlfirst{61.54}
& \hlfirst{65.38} & \hlfirst{92.31}
& \hlfirst{56.70} & \hlfirst{73.77} \\
\textit{Impro(\%)}
& \textit{+20.00} & \textit{+6.25}
& \textit{+15.00} & \textit{+14.27}
& \textit{+2.70} & \textit{+8.32}
& \textit{+4.18} & \textit{+0.00}
& \textit{+6.24} & \textit{+20.01}
& \textit{+17.49} & \textit{+14.18} \\
\bottomrule
\end{tabular}
\vspace{-0.2cm}
\captionsetup{font=footnotesize}
\caption{Performance comparison on \textsc{MCPMark} under the cross-agent transfer setting, using GPT-5.4 as the backbone. All baselines build their memory with \textit{SwitchAct} as the source agent, then transfer to two target agents: \textit{ReAct} and \textit{CodeAct}. The \textbf{best} and \underline{second best} results are marked in bold and with underline, respectively. \textit{Impro(\%)} is the relative gain of \mname{} over the strongest baseline.}
\label{tab:rq2_cross_agent}
\end{table*}
\begin{table*}[!ht]
\centering
\fontsize{7pt}{8pt}\selectfont
\setlength{\tabcolsep}{2.8pt}
\renewcommand{\arraystretch}{1.05}
\begin{tabular}{l|cc|cc|cc|cc|cc|cc}
\toprule
\multirow{2}{*}{\textbf{Method}}
& \multicolumn{2}{c|}{\textbf{Filesystem}}
& \multicolumn{2}{c|}{\textbf{GitHub}}
& \multicolumn{2}{c|}{\textbf{Notion}}
& \multicolumn{2}{c|}{\textbf{Playwright}}
& \multicolumn{2}{c|}{\textbf{PostgreSQL}}
& \multicolumn{2}{c}{\textbf{Overall}} \\
& pass@1$\uparrow$ & pass@4$\uparrow$
& pass@1$\uparrow$ & pass@4$\uparrow$
& pass@1$\uparrow$ & pass@4$\uparrow$
& pass@1$\uparrow$ & pass@4$\uparrow$
& pass@1$\uparrow$ & pass@4$\uparrow$
& pass@1$\uparrow$ & pass@4$\uparrow$ \\
\midrule
\rowcolor[HTML]{F0F6FF}
\mname{} (Full)
& \hlfirst{69.73} & \hlfirst{94.73} & \hlfirst{56.67} & \hlfirst{73.33} & \hlfirst{51.39} & \hlfirst{72.22} & \hlfirst{59.62} & \hlfirst{76.92} & \hlfirst{67.30} & \hlfirst{84.61} & \hlfirst{60.94} & \hlfirst{80.36} \\
\midrule
\rowcolor{gray!10}
\multicolumn{13}{c}{\textit{Brainstorm Mechanism}} \\
\midrule
w/o Boundary
& 65.79 & 89.47
& 55.00 & 66.67
& 48.61 & 66.67
& 53.85 & 69.23
& 63.46 & 76.92
& 57.34 & 73.79 \\
w/o Affordance
& 61.84 & 84.21
& 48.33 & 60.00
& 50.00 & 66.67
& 50.00 & 61.54
& 57.69 & 69.23
& 53.57 & 68.33 \\
\midrule
\rowcolor{gray!10}
\multicolumn{13}{c}{\textit{Memory Organization}} \\
\midrule
w/o $\mathcal{G}_{\mathsf{cap}}$
& 55.26 & 78.95
& 50.00 & 60.00
& 45.83 & 61.11
& 51.92 & 61.54
& 55.77 & 69.23
& 51.76 & 66.17 \\
w/o $\mathcal{G}_{\mathsf{strat}}$
& 68.42 & 94.73
& 53.33 & 66.67
& 48.61 & 66.67
& 57.69 & 76.92
& 65.38 & 76.92
& 58.69 & 76.38 \\
w/o Dynamic Traversal
& 67.11 & 89.47
& 51.67 & 66.67
& 50.00 & 66.67
& 55.77 & 69.23
& 61.54 & 76.92
& 57.22 & 73.79 \\
\bottomrule
\end{tabular}
\vspace{-0.2cm}
\captionsetup{font=footnotesize}
\caption{Ablation study on \textsc{MCPMark} under the same-environment split, using \textit{SwitchAct} with GPT-5.4 as the backbone.  The \textbf{best} results are marked in bold.}
\label{tab:rq3_ablation}
\vspace{-0.6cm}
\end{table*}
\section{Experiment}
Our experiments evaluate the effectiveness and generalization of \mname{}. Specifically, we ask
whether the provider-side tool memory can be built once, and then reused by different agents, models, and
environment instances. We organize the evaluation around four
questions: 
 (\textbf{RQ1}) How does \mname perform compared with tool-side optimization and agent-side memory baselines? 
(\textbf{RQ2}) Does the tool-side memory constructed by \mname generalize across environment instances and agent frameworks? 
(\textbf{RQ3}) How much does each key design choice in \mname{}
contribute to its performance gains?
(\textbf{RQ4}) What inference cost does \mname{} incur, and
how does its utility-cost trade-off compare with
baselines?

\subsection{Experimental Setup}
\label{sec:exp:setup}

\noindent\textbf{Benchmarks.}
To thoroughly examine the 
effectiveness of \mname{}, We evaluate on two MCP-based benchmarks covering eight services.
First, we use \textsc{MCPMark}~\citep{DBLP:journals/corr/abs-2509-24002},
which includes \textit{Filesystem}, \textit{GitHub}, 
\textit{Notion}, \textit{Playwright}, and \textit{PostgreSQL}.
Second, we use \textsc{MCP-Universe}~\citep{DBLP:journals/corr/abs-2508-14704},
from which we select \textit{Location Navigation}, 
\textit{Financial Analysis}, and \textit{Web Searching}. We exclude
\textsc{MCP-Universe} domains that overlap with \textsc{MCPMark}
so that the combined evaluation covers a broader range of tool
interfaces and task types. Dataset details and split construction
are provided in Appendix~\ref{app:dataset}.

\noindent\textbf{Baselines.}
We compare \mname{} against two complementary families of methods. The first  targets 
\textit{tool-side optimization}, which improve the
descriptions, examples, or instructions of the tools, including EasyTool~\citep{yuan2025easytool},
Play2Prompt~\citep{fang2025play2prompt}, and
ToolOptimal~\citep{wu2025joint}. The second family are
\emph{agent-side memory} methods, which summarize past trajectories
into reusable experience associated with a particular agent, including ExpeL~\citep{zhao2024expel},
A-MEM~\citep{xu2026mem}, and ACE~\citep{zhang2025agentic}. We also include \emph{Vanilla}, the agent with raw MCP tool specifications and no augmentation. Full details are in Appendix~\ref{app:baseline}. 

\noindent\textbf{Agents and LLM Backbones.}
We instantiate three representative tool-using agent frameworks (details in Appendix~\ref{app:agent-setup}). \textbf{ReAct}~\citep{yao2022react} issues each action as a
single function call. \textbf{CodeAct}~\citep{wang2024executable}
issues actions as executable Python snippets. \textbf{SwitchAct}
is a hybrid agent that we implement, exposing both interfaces to
the LLM and allowing it to choose at each
step. These frameworks cover both function-calling and
code-based tool-use paradigms. Unless otherwise stated, we use
SwitchAct as the default inference agent. We evaluate three LLM backbones from two model
families: GPT-5.4, GPT-5.4-mini, and Grok-4 (implementation details in
Appendix~\ref{app:llm-backbone}).

\noindent\textbf{Metrics and Protocol.}
We report pass@1 and pass@4 following the benchmarks' setting, and total inference tokens to measure cost.
For methods that construct memory, the memory is built only on
the designated training split and then frozen before evaluation.
For \mname{}, the resulting tool memory is associated with the tool rather than with a specific downstream agent. This
allows us to evaluate whether the same tool-side memory can be
reused across agents and environments without retraining or
additional trajectory accumulation.

\subsection{RQ1: Effectiveness of \mname{}}
\label{sec:exp:rq1}

We first evaluate on the \emph{same-environment split}
(Appendix~\ref{app:split}), where training and test tasks share
the same environment instances. This setting isolates the effect
of the memory mechanism, since no environment-instance
shift is introduced between memory construction and evaluation.
\Cref{tab:rq1_mcpmark_gpt54_gpt54mini,tab:rq1_mcpmark_grok}
report results on \textsc{MCPMark} across three LLM backbones,
and \Cref{tab:rq1_mcpuniverse} reports results on
\textsc{MCP-Universe}.
\mname{} outperforms the strongest baseline on every service and with every backbone on \textsc{MCPMark}. Relative to the best baseline, \mname{} improves pass@1/pass@4 by
$8.15\%$/$10.48\%$ with GPT-5.4,
$10.06\%$/$18.61\%$ with GPT-5.4-mini, and
$21.61\%$/$16.11\%$ with Grok-4. The same trend extends
to \textsc{MCP-Universe} results. These results show
that the gains are not tied to a single benchmark, or
model backbone. Instead, they support the central premise of
\mname{}: execution-verified, tool-bound knowledge can improve
downstream tool use even when the agent itself is unchanged.
The relative gains are larger for GPT-5.4-mini and
Grok-4 than for GPT-5.4, which suggests that a
provider-side tool memory does more than improve already
strong agents. It can also compensate for weaker backbones by supplying reusable knowledge about
tool behavior at inference time.



\subsection{RQ2: Generalization of \mname{}}
\label{sec:exp:rq2}

The defining advantage of \mname{} is not only that it improves
one agent, but that it places tool-use knowledge at the provider
side, where the same tool memory can serve many downstream consumers.
We therefore evaluate its generalization under two deployment shifts:
cross-environment transfer, where the tool service is the same
but the underlying environment instance changes, and cross-agent transfer,
where the tool memory is built with one agent and consumed by others.

\noindent\textbf{Across Environment Instances.} 
In this setting, training and test tasks use
disjoint environment instances while keeping the tool set fixed
(Appendix~\ref{app:split}). For example, the tool memory may be built
from one repository, workspace, or database instance, and evaluated on another. This setting tests whether a method learns
generalized tool behaviors rather than memorizing environment-specific
artifacts.
\Cref{tab:rq2_cross_environment} shows that \mname{} achieves
the best performance and outperforms the strongest
baseline in every service, with relative gains of
$24.16\%$/$16.22\%$ on pass@1/pass@4. In contrast, the same
shift degrades many tool-side and agent-side baselines that
fall $5$--$7$ percentage points below \emph{Vanilla} on overall
pass@1. This indicates that these methods are absorbing
environment-specific patterns that cannot transfer. By organizing
memory around tool capabilities and validating entries through
execution, \mname{} captures knowledge that remains useful across environment instances of the same service.

\noindent\textbf{Across Agent Frameworks.}
We test whether different agent frameworks can reuse the tool memory. Each method first constructs memory with SwitchAct and then transfers it to ReAct, which uses function calls, and CodeAct,
which uses executable Python snippets, respectively. This setting reflects a
realistic deployment: the provider maintains one
tool-side memory, while heterogeneous downstream agents query the same memory instance.
\mname{} achieves the largest gains on both target agents as shown in \Cref{tab:rq2_cross_agent}: $+16.25\%$/$+14.27\%$ on
ReAct and $+17.49\%$/$+14.18\%$ on CodeAct, measured by overall
pass@1/pass@4. In contrast, agent-side memory baselines often
fall below \emph{Vanilla} after transfer. For example, A-MEM loses
$8.78\%$ on ReAct overall pass@1 and $26.92\%$
percentage points on CodeAct PostgreSQL pass@1. These failures
highlight a limitation of agent-bound memory: the stored lessons encode the workflow, action format, and failure modes of the specific agent that generated them, while \mname{} avoids this coupling by storing generalized knowledge at the tool level.

\subsection{RQ3: Ablation Study}
\label{sec:exp:rq3}

\Cref{tab:rq3_ablation} presents an ablation of \mname{} by 
isolating each design choice in the brainstorm mechanism and the 
memory organization.
Removing any component hurts performance, showing that the gains come from coupling a tool-centered memory structure with frontier-guided execution
probing.
The largest drop comes from removing the Tool-Capability Graph: \textit{w/o $\mathcal{G}_{\mathsf{cap}}$} decreases pass@1/pass@4 by
$9.18\%$/$14.19\%$. This directly supports the design principle
of \mname{}: reusable tool knowledge should be centered at the
tool and its capabilities, not at the idiosyncratic trajectory of the agent that discovered it.
During the brainstorm phase, removing affordance exploration
causes larger degradation than removing boundary exploration.
Specifically, \textit{w/o Affordance} reduces pass@1/pass@4 by
$7.37\%$/$12.03\%$, whereas \textit{w/o Boundary} reduces by
$3.60\%$/$6.57\%$. This suggests that affordance exploration is
the primary source of positive tool-use knowledge, while boundary
exploration provides complementary information about where tools
fail or behave unreliably. Removing \textit{Dynamic Traversal}
and the \textit{Tool-Strategy Graph} also produces consistent drops,
showing that adaptive exploration and higher-level organization
further improve the atlas beyond the basic tool-bound memory
structure. Appendix~\ref{app:ablation_frontier} further shows that Frontier
Exploration avoids the plateau of same-budget seed-task sampling, suggesting \mname{} constructs less redundant tool memory.

\subsection{RQ4 Inference and Utility Cost Trade-off}
\label{sec:exp:rq4}
\begin{figure}[t]
  \centering
  \includegraphics[width=0.45\textwidth]{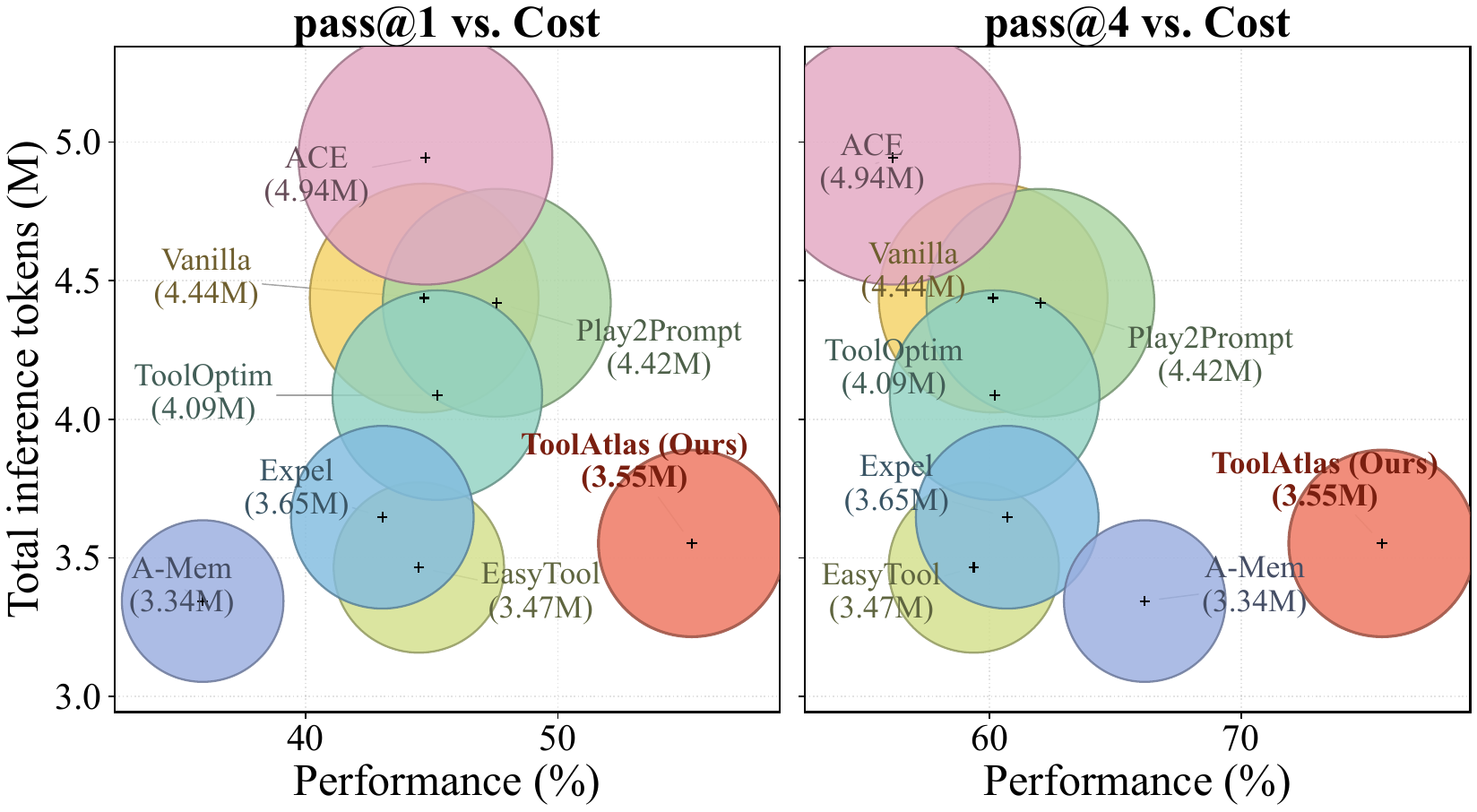}
  \captionsetup{font=footnotesize} \caption{Utility--cost trade-off on \textsc{MCPMark} under the cross-agent transfer setting, using GPT-5.4 as the backbone. Total inference tokens are annotated per method.}
  \vspace{-4mm}
  \label{fig:rq4_cost}
\end{figure}

Under the cross-agent transfer setting (\Cref{sec:exp:rq2}), \Cref{fig:rq4_cost} plots pass@1/pass@4 against inference tokens on \textsc{MCPMark}.
\mname{} lies on the Pareto frontier in both, reaching highest task success with 3.55M tokens, below Vanilla's 4.44M. The saving comes from avoiding tool rediscovery, since provider-side memory supplies execution-verified affordances and boundaries upfront, so rollout savings exceed the prompt-side cost. Baselines miss this balance. Tool-side optimization is compact but less accurate, while agent-side memory adds prompt tokens without matching gains. \mname{} avoids both ends by storing execution-verified, agent-neutral tool knowledge with the tool, giving the receiving agent a generalizable capability map instead of forcing re-exploration.


\section{Conclusion}
Tool memory should be maintained provider-side rather than inside individual agents. We realize this view in \mname{}, which charts each MCP server as a three-layer tool memory graph, expands it through brainstorm-driven exploration of tool affordances and boundaries, and surfaces task-conditioned guidance at inference. Across eight services and two benchmarks, \mname{} outperforms tool-side and agent-side baselines, transfers across environment instances and agent frameworks without retraining, and reduces end-to-end inference cost. Provider-side memory offers a practical paradigm for shared tool servers, amortizing one offline effort across many downstream agents.

\section{Limitations}
Although \mname{} demonstrates strong effectiveness and transferability
on MCP-based benchmarks, several limitations remain. Our
evaluation focuses on eight services from two MCP benchmarks and three
LLM backbones. This setting covers diverse tool types, but it does not exhaust
the full space of real-world tool servers. In particular, we do not
evaluate tools in highly specialized domain constraints or rapidly
changing APIs. Extending the evaluation to broader production tool ecosystems is an important direction for future work. In addition, our  study emphasizes task success, transfer across environment instances and agent frameworks, component ablations, and inference cost. We do not fully account for the total lifecycle of provider-side tool-memory, especially regarding to re-verification, storage, and governance. Tool behavior may change as APIs are updated, permissions are modified, or backend data schemas evolve. A practical deployment of \mname{} would therefore require dedicated verification and refresh mechanisms to ensure that stored tool knowledge remains accurate over time.

\section{Ethical Considerations}
All experiments in this study were conducted on publicly available MCP-based benchmarks, including \textsc{MCPMark} and \textsc{MCP-Universe}, in compliance with their respective licenses and usage terms. When external services or accounts were required to instantiate benchmark environments, we used researcher-controlled accounts and synthetic or benchmark-provided data only. No third-party user accounts, private user data, or production workloads were involved. This work did not involve human or animal subjects, and no personally identifiable information was collected, or used. We further note that a provider-side tool memory could, in principle, be poisoned or exploited if constructed from untrusted execution traces, and any deployment should pair it with verification and access-control mechanisms before serving downstream agents.

\bibliography{custom}

@article{xiao2025toolmem,
  author       = {Yunzhong Xiao and
                  Yangmin Li and
                  Hewei Wang and
                  Yunlong Tang and
                  Zora Zhiruo Wang},
  title        = {ToolMem: Enhancing Multimodal Agents with Learnable Tool Capability
                  Memory},
  journal      = {CoRR},
  volume       = {abs/2510.06664},
  year         = {2025},
  url          = {https://doi.org/10.48550/arXiv.2510.06664},
  doi          = {10.48550/ARXIV.2510.06664},
  eprinttype   = {arXiv},
  eprint       = {2510.06664},
  bibsource    = {dblp computer science bibliography, https://dblp.org}
}

@article{hsieh2023tool,
  author       = {Cheng{-}Yu Hsieh and
                  Si{-}An Chen and
                  Chun{-}Liang Li and
                  Yasuhisa Fujii and
                  Alexander Ratner and
                  Chen{-}Yu Lee and
                  Ranjay Krishna and
                  Tomas Pfister},
  title        = {Tool Documentation Enables Zero-Shot Tool-Usage with Large Language
                  Models},
  journal      = {CoRR},
  volume       = {abs/2308.00675},
  year         = {2023},
  url          = {https://doi.org/10.48550/arXiv.2308.00675},
  doi          = {10.48550/ARXIV.2308.00675},
  eprinttype   = {arXiv},
  eprint       = {2308.00675},
  bibsource    = {dblp computer science bibliography, https://dblp.org}
}

@inproceedings{qu2025exploration,
 author       = {Changle Qu and
                  Sunhao Dai and
                  Xiaochi Wei and
                  Hengyi Cai and
                  Shuaiqiang Wang and
                  Dawei Yin and
                  Jun Xu and
                  Ji{-}Rong Wen},
  title        = {From Exploration to Mastery: Enabling LLMs to Master Tools via Self-Driven
                  Interactions},
  booktitle    = {The Thirteenth International Conference on Learning Representations,
                  {ICLR} 2025, Singapore, April 24-28, 2025},
  publisher    = {OpenReview.net},
  year         = {2025},
  url          = {https://openreview.net/forum?id=QKBu1BOAwd},
  bibsource    = {dblp computer science bibliography, https://dblp.org}
}

@inproceedings{chen2024reinvoke,
  author       = {Yanfei Chen and
                  Jinsung Yoon and
                  Devendra Singh Sachan and
                  Qingze Wang and
                  Vincent Cohen{-}Addad and
                  MohammadHossein Bateni and
                  Chen{-}Yu Lee and
                  Tomas Pfister},
  editor       = {Yaser Al{-}Onaizan and
                  Mohit Bansal and
                  Yun{-}Nung Chen},
  title        = {Re-Invoke: Tool Invocation Rewriting for Zero-Shot Tool Retrieval},
  booktitle    = {Findings of the Association for Computational Linguistics: {EMNLP}
                  2024, Miami, Florida, USA, November 12-16, 2024},
  series       = {Findings of {ACL}},
  pages        = {4705--4726},
  publisher    = {Association for Computational Linguistics},
  year         = {2024},
  url          = {https://doi.org/10.18653/v1/2024.findings-emnlp.270},
  doi          = {10.18653/V1/2024.FINDINGS-EMNLP.270},
  bibsource    = {dblp computer science bibliography, https://dblp.org}
}

@inproceedings{shi2025toolret,
  author       = {Zhengliang Shi and
                  Yuhan Wang and
                  Lingyong Yan and
                  Pengjie Ren and
                  Shuaiqiang Wang and
                  Dawei Yin and
                  Zhaochun Ren},
  editor       = {Wanxiang Che and
                  Joyce Nabende and
                  Ekaterina Shutova and
                  Mohammad Taher Pilehvar},
  title        = {Retrieval Models Aren't Tool-Savvy: Benchmarking Tool Retrieval for
                  Large Language Models},
  booktitle    = {Findings of the Association for Computational Linguistics, {ACL} 2025,
                  Vienna, Austria, July 27 - August 1, 2025},
  series       = {Findings of {ACL}},
  pages        = {24497--24524},
  publisher    = {Association for Computational Linguistics},
  year         = {2025},
  url          = {https://aclanthology.org/2025.findings-acl.1258/},
  bibsource    = {dblp computer science bibliography, https://dblp.org}
}

@InProceedings{du2024anytool,
  author       = {Yu Du and
                  Fangyun Wei and
                  Hongyang Zhang},
  editor       = {Ruslan Salakhutdinov and
                  Zico Kolter and
                  Katherine A. Heller and
                  Adrian Weller and
                  Nuria Oliver and
                  Jonathan Scarlett and
                  Felix Berkenkamp},
  title        = {AnyTool: Self-Reflective, Hierarchical Agents for Large-Scale {API}
                  Calls},
  booktitle    = {Forty-first International Conference on Machine Learning, {ICML} 2024,
                  Vienna, Austria, July 21-27, 2024},
  series       = {Proceedings of Machine Learning Research},
  pages        = {11812--11829},
  publisher    = {{PMLR} / OpenReview.net},
  year         = {2024},
  url          = {https://proceedings.mlr.press/v235/du24h.html},
  bibsource    = {dblp computer science bibliography, https://dblp.org}
}

@article{liu2025toolscope,
  author       = {Marianne Menglin Liu and
                  Daniel Garcia and
                  Fjona Parllaku and
                  Vikas Upadhyay and
                  Syed Fahad Allam Shah and
                  Dan Roth},
  title        = {ToolScope: Enhancing {LLM} Agent Tool Use through Tool Merging and
                  Context-Aware Filtering},
  journal      = {CoRR},
  volume       = {abs/2510.20036},
  year         = {2025},
  url          = {https://doi.org/10.48550/arXiv.2510.20036},
  doi          = {10.48550/ARXIV.2510.20036},
  eprinttype   = {arXiv},
  eprint       = {2510.20036},
  bibsource    = {dblp computer science bibliography, https://dblp.org}
}

@inproceedings{zhang2025toolexpnet,
  author       = {Zijing Zhang and
                  Zhanpeng Chen and
                  He Zhu and
                  Ziyang Chen and
                  Nan Du and
                  Xiaolong Li},
  editor       = {Wanxiang Che and
                  Joyce Nabende and
                  Ekaterina Shutova and
                  Mohammad Taher Pilehvar},
  title        = {ToolExpNet: Optimizing Multi-Tool Selection in LLMs with Similarity
                  and Dependency-Aware Experience Networks},
  booktitle    = {Findings of the Association for Computational Linguistics, {ACL} 2025,
                  Vienna, Austria, July 27 - August 1, 2025},
  series       = {Findings of {ACL}},
  pages        = {15706--15722},
  publisher    = {Association for Computational Linguistics},
  year         = {2025},
  url          = {https://aclanthology.org/2025.findings-acl.811/},
  bibsource    = {dblp computer science bibliography, https://dblp.org}
}

@inproceedings{cai2024large,
  author       = {Tianle Cai and
                  Xuezhi Wang and
                  Tengyu Ma and
                  Xinyun Chen and
                  Denny Zhou},
  title        = {Large Language Models as Tool Makers},
  booktitle    = {The Twelfth International Conference on Learning Representations,
                  {ICLR} 2024, Vienna, Austria, May 7-11, 2024},
  publisher    = {OpenReview.net},
  year         = {2024},
  url          = {https://openreview.net/forum?id=qV83K9d5WB},
  bibsource    = {dblp computer science bibliography, https://dblp.org}
}

@inproceedings{qian2023creator,
  author       = {Cheng Qian and
                  Chi Han and
                  Yi Ren Fung and
                  Yujia Qin and
                  Zhiyuan Liu and
                  Heng Ji},
  editor       = {Houda Bouamor and
                  Juan Pino and
                  Kalika Bali},
  title        = {{CREATOR:} Tool Creation for Disentangling Abstract and Concrete Reasoning
                  of Large Language Models},
  booktitle    = {Findings of the Association for Computational Linguistics: {EMNLP}
                  2023, Singapore, December 6-10, 2023},
  series       = {Findings of {ACL}},
  pages        = {6922--6939},
  publisher    = {Association for Computational Linguistics},
  year         = {2023},
  url          = {https://doi.org/10.18653/v1/2023.findings-emnlp.462},
  doi          = {10.18653/V1/2023.FINDINGS-EMNLP.462},
  bibsource    = {dblp computer science bibliography, https://dblp.org}
}

@inproceedings{yuan2024craft,
  author       = {Lifan Yuan and
                  Yangyi Chen and
                  Xingyao Wang and
                  Yi Fung and
                  Hao Peng and
                  Heng Ji},
  title        = {{CRAFT:} Customizing LLMs by Creating and Retrieving from Specialized
                  Toolsets},
  booktitle    = {The Twelfth International Conference on Learning Representations,
                  {ICLR} 2024, Vienna, Austria, May 7-11, 2024},
  publisher    = {OpenReview.net},
  year         = {2024},
  url          = {https://openreview.net/forum?id=G0vdDSt9XM},
  bibsource    = {dblp computer science bibliography, https://dblp.org}
}

@inproceedings{wang2024executable,
  author       = {Xingyao Wang and
                  Yangyi Chen and
                  Lifan Yuan and
                  Yizhe Zhang and
                  Yunzhu Li and
                  Hao Peng and
                  Heng Ji},
  editor       = {Ruslan Salakhutdinov and
                  Zico Kolter and
                  Katherine A. Heller and
                  Adrian Weller and
                  Nuria Oliver and
                  Jonathan Scarlett and
                  Felix Berkenkamp},
  title        = {Executable Code Actions Elicit Better {LLM} Agents},
  booktitle    = {Forty-first International Conference on Machine Learning, {ICML} 2024,
                  Vienna, Austria, July 21-27, 2024},
  series       = {Proceedings of Machine Learning Research},
  pages        = {50208--50232},
  publisher    = {{PMLR} / OpenReview.net},
  year         = {2024},
  url          = {https://proceedings.mlr.press/v235/wang24h.html},
  bibsource    = {dblp computer science bibliography, https://dblp.org}
}

@article{yao2022react,
  author       = {Shunyu Yao and
                  Jeffrey Zhao and
                  Dian Yu and
                  Nan Du and
                  Izhak Shafran and
                  Karthik R. Narasimhan and
                  Yuan Cao},
  title        = {ReAct: Synergizing Reasoning and Acting in Language Models},
  booktitle    = {The Eleventh International Conference on Learning Representations,
                  {ICLR} 2023, Kigali, Rwanda, May 1-5, 2023},
  publisher    = {OpenReview.net},
  year         = {2023},
  url          = {https://openreview.net/forum?id=WE\_vluYUL-X},
  bibsource    = {dblp computer science bibliography, https://dblp.org}
}

@article{wang2023voyager,
  author       = {Guanzhi Wang and
                  Yuqi Xie and
                  Yunfan Jiang and
                  Ajay Mandlekar and
                  Chaowei Xiao and
                  Yuke Zhu and
                  Linxi Fan and
                  Anima Anandkumar},
  title        = {Voyager: An Open-Ended Embodied Agent with Large Language Models},
  journal      = {Trans. Mach. Learn. Res.},
  volume       = {2024},
  year         = {2024},
  url          = {https://openreview.net/forum?id=ehfRiF0R3a},
  bibsource    = {dblp computer science bibliography, https://dblp.org}
}

@article{ouyang2025reasoningbank,
  author       = {Siru Ouyang and
                  Jun Yan and
                  I{-}Hung Hsu and
                  Yanfei Chen and
                  Ke Jiang and
                  Zifeng Wang and
                  Rujun Han and
                  Long T. Le and
                  Samira Daruki and
                  Xiangru Tang and
                  Vishy Tirumalashetty and
                  George Lee and
                  Mahsan Rofouei and
                  Hangfei Lin and
                  Jiawei Han and
                  Chen{-}Yu Lee and
                  Tomas Pfister},
  title        = {ReasoningBank: Scaling Agent Self-Evolving with Reasoning Memory},
  journal      = {CoRR},
  volume       = {abs/2509.25140},
  year         = {2025},
  url          = {https://doi.org/10.48550/arXiv.2509.25140},
  doi          = {10.48550/ARXIV.2509.25140},
  eprinttype   = {arXiv},
  eprint       = {2509.25140},
  bibsource    = {dblp computer science bibliography, https://dblp.org}
}

@article{zhang2025agentic,
  author       = {Qizheng Zhang and
                  Changran Hu and
                  Shubhangi Upasani and
                  Boyuan Ma and
                  Fenglu Hong and
                  Vamsidhar Kamanuru and
                  Jay Rainton and
                  Chen Wu and
                  Mengmeng Ji and
                  Hanchen Li and
                  Urmish Thakker and
                  James Zou and
                  Kunle Olukotun},
  title        = {Agentic Context Engineering: Evolving Contexts for Self-Improving
                  Language Models},
  journal      = {CoRR},
  volume       = {abs/2510.04618},
  year         = {2025},
  url          = {https://doi.org/10.48550/arXiv.2510.04618},
  doi          = {10.48550/ARXIV.2510.04618},
  eprinttype   = {arXiv},
  eprint       = {2510.04618},
  bibsource    = {dblp computer science bibliography, https://dblp.org}
}

@article{xu2026mem,
  author       = {Wujiang Xu and
                  Zujie Liang and
                  Kai Mei and
                  Hang Gao and
                  Juntao Tan and
                  Yongfeng Zhang},
  title        = {{A-MEM:} Agentic Memory for {LLM} Agents},
  journal      = {CoRR},
  volume       = {abs/2502.12110},
  year         = {2025},
  url          = {https://doi.org/10.48550/arXiv.2502.12110},
  doi          = {10.48550/ARXIV.2502.12110},
  eprinttype   = {arXiv},
  eprint       = {2502.12110},
  bibsource    = {dblp computer science bibliography, https://dblp.org}
}

@inproceedings{zhao2024expel,
  author       = {Andrew Zhao and
                  Daniel Huang and
                  Quentin Xu and
                  Matthieu Lin and
                  Yong{-}Jin Liu and
                  Gao Huang},
  editor       = {Michael J. Wooldridge and
                  Jennifer G. Dy and
                  Sriraam Natarajan},
  title        = {ExpeL: {LLM} Agents Are Experiential Learners},
  booktitle    = {Thirty-Eighth {AAAI} Conference on Artificial Intelligence, {AAAI}
                  2024, Thirty-Sixth Conference on Innovative Applications of Artificial
                  Intelligence, {IAAI} 2024, Fourteenth Symposium on Educational Advances
                  in Artificial Intelligence, {EAAI} 2014, February 20-27, 2024, Vancouver,
                  Canada},
  pages        = {19632--19642},
  publisher    = {{AAAI} Press},
  year         = {2024},
  url          = {https://doi.org/10.1609/aaai.v38i17.29936},
  doi          = {10.1609/AAAI.V38I17.29936},
  bibsource    = {dblp computer science bibliography, https://dblp.org}
}

@inproceedings{qin2024toolllm,
author       = {Yujia Qin and
                  Shihao Liang and
                  Yining Ye and
                  Kunlun Zhu and
                  Lan Yan and
                  Yaxi Lu and
                  Yankai Lin and
                  Xin Cong and
                  Xiangru Tang and
                  Bill Qian and
                  Sihan Zhao and
                  Lauren Hong and
                  Runchu Tian and
                  Ruobing Xie and
                  Jie Zhou and
                  Mark Gerstein and
                  Dahai Li and
                  Zhiyuan Liu and
                  Maosong Sun},
  title        = {ToolLLM: Facilitating Large Language Models to Master 16000+ Real-world
                  APIs},
  booktitle    = {The Twelfth International Conference on Learning Representations,
                  {ICLR} 2024, Vienna, Austria, May 7-11, 2024},
  publisher    = {OpenReview.net},
  year         = {2024},
  url          = {https://openreview.net/forum?id=dHng2O0Jjr},
  bibsource    = {dblp computer science bibliography, https://dblp.org}
}

@article{schick2023toolformer,
  author       = {Timo Schick and
                  Jane Dwivedi{-}Yu and
                  Roberto Dess{\`{\i}} and
                  Roberta Raileanu and
                  Maria Lomeli and
                  Eric Hambro and
                  Luke Zettlemoyer and
                  Nicola Cancedda and
                  Thomas Scialom},
  editor       = {Alice Oh and
                  Tristan Naumann and
                  Amir Globerson and
                  Kate Saenko and
                  Moritz Hardt and
                  Sergey Levine},
  title        = {Toolformer: Language Models Can Teach Themselves to Use Tools},
  booktitle    = {Advances in Neural Information Processing Systems 36: Annual Conference
                  on Neural Information Processing Systems 2023, NeurIPS 2023, New Orleans,
                  LA, USA, December 10 - 16, 2023},
  year         = {2023},
  url          = {http://papers.nips.cc/paper\_files/paper/2023/hash/d842425e4bf79ba039352da0f658a906-Abstract-Conference.html},
  bibsource    = {dblp computer science bibliography, https://dblp.org}
}

@inproceedings{liu2025toolace,
  author       = {Weiwen Liu and
                  Xu Huang and
                  Xingshan Zeng and
                  Xinlong Hao and
                  Shuai Yu and
                  Dexun Li and
                  Shuai Wang and
                  Weinan Gan and
                  Zhengying Liu and
                  Yuanqing Yu and
                  Zezhong Wang and
                  Yuxian Wang and
                  Wu Ning and
                  Yutai Hou and
                  Bin Wang and
                  Chuhan Wu and
                  Xinzhi Wang and
                  Yong Liu and
                  Yasheng Wang and
                  Duyu Tang and
                  Dandan Tu and
                  Lifeng Shang and
                  Xin Jiang and
                  Ruiming Tang and
                  Defu Lian and
                  Qun Liu and
                  Enhong Chen},
  title        = {ToolACE: Winning the Points of {LLM} Function Calling},
  booktitle    = {The Thirteenth International Conference on Learning Representations,
                  {ICLR} 2025, Singapore, April 24-28, 2025},
  publisher    = {OpenReview.net},
  year         = {2025},
  url          = {https://openreview.net/forum?id=8EB8k6DdCU},
  bibsource    = {dblp computer science bibliography, https://dblp.org}
}

@inproceedings{wu2025joint,
  author       = {Bin Wu and
                  Edgar Meij and
                  Emine Yilmaz},
  editor       = {Wanxiang Che and
                  Joyce Nabende and
                  Ekaterina Shutova and
                  Mohammad Taher Pilehvar},
  title        = {A Joint Optimization Framework for Enhancing Efficiency of Tool Utilization
                  in {LLM} Agents},
  booktitle    = {Findings of the Association for Computational Linguistics, {ACL} 2025,
                  Vienna, Austria, July 27 - August 1, 2025},
  series       = {Findings of {ACL}},
  pages        = {22361--22373},
  publisher    = {Association for Computational Linguistics},
  year         = {2025},
  url          = {https://aclanthology.org/2025.findings-acl.1149/},
  bibsource    = {dblp computer science bibliography, https://dblp.org}
}

@inproceedings{fang2025play2prompt,
  author       = {Wei Fang and
                  Yang Zhang and
                  Kaizhi Qian and
                  James R. Glass and
                  Yada Zhu},
  editor       = {Wanxiang Che and
                  Joyce Nabende and
                  Ekaterina Shutova and
                  Mohammad Taher Pilehvar},
  title        = {{PLAY2PROMPT:} Zero-shot Tool Instruction Optimization for {LLM} Agents
                  via Tool Play},
  booktitle    = {Findings of the Association for Computational Linguistics, {ACL} 2025,
                  Vienna, Austria, July 27 - August 1, 2025},
  series       = {Findings of {ACL}},
  pages        = {26274--26290},
  publisher    = {Association for Computational Linguistics},
  year         = {2025},
  url          = {https://aclanthology.org/2025.findings-acl.1347/},
  bibsource    = {dblp computer science bibliography, https://dblp.org}
}

@inproceedings{yuan2025easytool,
  author       = {Siyu Yuan and
                  Kaitao Song and
                  Jiangjie Chen and
                  Xu Tan and
                  Yongliang Shen and
                  Kan Ren and
                  Dongsheng Li and
                  Deqing Yang},
  editor       = {Luis Chiruzzo and
                  Alan Ritter and
                  Lu Wang},
  title        = {{EASYTOOL:} Enhancing LLM-based Agents with Concise Tool Instruction},
  booktitle    = {Proceedings of the 2025 Conference of the Nations of the Americas
                  Chapter of the Association for Computational Linguistics: Human Language
                  Technologies, {NAACL} 2025 - Volume 1: Long Papers, Albuquerque, New
                  Mexico, USA, April 29 - May 4, 2025},
  pages        = {951--972},
  publisher    = {Association for Computational Linguistics},
  year         = {2025},
  url          = {https://doi.org/10.18653/v1/2025.naacl-long.44},
  doi          = {10.18653/V1/2025.NAACL-LONG.44},
  bibsource    = {dblp computer science bibliography, https://dblp.org}
}

@article{DBLP:journals/corr/abs-2509-24002,
  author       = {Zijian Wu and
                  Xiangyan Liu and
                  Xinyuan Zhang and
                  Lingjun Chen and
                  Fanqing Meng and
                  Lingxiao Du and
                  Yiran Zhao and
                  Fanshi Zhang and
                  Yaoqi Ye and
                  Jiawei Wang and
                  Zirui Wang and
                  Jinjie Ni and
                  Yufan Yang and
                  Arvin Xu and
                  Michael Qizhe Shieh},
  title        = {MCPMark: {A} Benchmark for Stress-Testing Realistic and Comprehensive
                  {MCP} Use},
  journal      = {CoRR},
  volume       = {abs/2509.24002},
  year         = {2025},
  url          = {https://doi.org/10.48550/arXiv.2509.24002},
  doi          = {10.48550/ARXIV.2509.24002},
  eprinttype   = {arXiv},
  eprint       = {2509.24002},
  bibsource    = {dblp computer science bibliography, https://dblp.org}
}

@article{DBLP:journals/corr/abs-2508-14704,
  author       = {Ziyang Luo and
                  Zhiqi Shen and
                  Wenzhuo Yang and
                  Zirui Zhao and
                  Prathyusha Jwalapuram and
                  Amrita Saha and
                  Doyen Sahoo and
                  Silvio Savarese and
                  Caiming Xiong and
                  Junnan Li},
  title        = {MCP-Universe: Benchmarking Large Language Models with Real-World Model
                  Context Protocol Servers},
  journal      = {CoRR},
  volume       = {abs/2508.14704},
  year         = {2025},
  url          = {https://doi.org/10.48550/arXiv.2508.14704},
  doi          = {10.48550/ARXIV.2508.14704},
  eprinttype   = {arXiv},
  eprint       = {2508.14704},
  bibsource    = {dblp computer science bibliography, https://dblp.org}
}

@misc{mcp2026intro,
  author       = {{Model Context Protocol}},
  title        = {What is the Model Context Protocol (MCP)?},
  year         = {2026},
  howpublished = {\url{https://modelcontextprotocol.io/docs/getting-started/intro}},
  note         = {Accessed: 05-01-2026}
}

@misc{openai2026mcpconnectors,
  author       = {{OpenAI}},
  title        = {MCP and Connectors},
  year         = {2026},
  howpublished = {\url{https://developers.openai.com/api/docs/guides/tools-connectors-mcp}},
  note         = {Accessed: 05-01-2026}
}

@misc{github2026mcpserver,
  author       = {{GitHub}},
  title        = {Using the GitHub MCP Server in your IDE},
  year         = {2026},
  howpublished = {\url{https://docs.github.com/en/copilot/how-tos/provide-context/use-mcp-in-your-ide/use-the-github-mcp-server}},
  note         = {Accessed: 05-01-2026}
}

@misc{stripe2026mcp,
  author       = {{Stripe}},
  title        = {Model Context Protocol (MCP)},
  year         = {2026},
  howpublished = {\url{https://docs.stripe.com/mcp}},
  note         = {Accessed: 05-01-2026}
}

@inproceedings{DBLP:conf/nips/PatilZ0G24,
  author       = {Shishir G. Patil and
                  Tianjun Zhang and
                  Xin Wang and
                  Joseph E. Gonzalez},
  editor       = {Amir Globersons and
                  Lester Mackey and
                  Danielle Belgrave and
                  Angela Fan and
                  Ulrich Paquet and
                  Jakub M. Tomczak and
                  Cheng Zhang},
  title        = {Gorilla: Large Language Model Connected with Massive APIs},
  booktitle    = {Advances in Neural Information Processing Systems 38: Annual Conference
                  on Neural Information Processing Systems 2024, NeurIPS 2024, Vancouver,
                  BC, Canada, December 10 - 15, 2024},
  year         = {2024},
  url          = {http://papers.nips.cc/paper\_files/paper/2024/hash/e4c61f578ff07830f5c37378dd3ecb0d-Abstract-Conference.html},
  bibsource    = {dblp computer science bibliography, https://dblp.org}
}

@inproceedings{DBLP:conf/emnlp/LiZ000YLHL23,
  author       = {Minghao Li and
                  Yingxiu Zhao and
                  Bowen Yu and
                  Feifan Song and
                  Hangyu Li and
                  Haiyang Yu and
                  Zhoujun Li and
                  Fei Huang and
                  Yongbin Li},
  editor       = {Houda Bouamor and
                  Juan Pino and
                  Kalika Bali},
  title        = {API-Bank: {A} Comprehensive Benchmark for Tool-Augmented LLMs},
  booktitle    = {Proceedings of the 2023 Conference on Empirical Methods in Natural
                  Language Processing, {EMNLP} 2023, Singapore, December 6-10, 2023},
  pages        = {3102--3116},
  publisher    = {Association for Computational Linguistics},
  year         = {2023},
  url          = {https://doi.org/10.18653/v1/2023.emnlp-main.187},
  doi          = {10.18653/V1/2023.EMNLP-MAIN.187},
  bibsource    = {dblp computer science bibliography, https://dblp.org}
}

\appendix


\newpage

\section{Experimental Details}\label{app:exp-details}
\subsection{Dataset Descriptions}\label{app:dataset}

We evaluate \mname{} on two MCP-based benchmarks that test 
tool-using capabilities across diverse real-world environments. We 
first introduce the terminology used to describe benchmark 
organization:
\begin{itemize}[leftmargin=*,nosep]
    \item \textbf{MCP service} denotes a tool platform (\textit{e.g.}, \textit{Filesystem}, \textit{GitHub}). The 
    tool set of each service is fixed and shared across all its 
    environment instances and tasks.
    
    \item \textbf{Environment instance} denotes one of multiple 
    concrete, pre-populated realizations of an MCP service 
    (\textit{e.g.}, a specific repository, workspace, database, or 
    directory tree), each reusable across multiple tasks.
    
    \item \textbf{Task} denotes a single evaluation unit defined over 
    an environment instance, comprising a natural-language 
    instruction, a programmatic verifier, and a task-specific starting 
    snapshot of that instance.
\end{itemize}

We adopt two benchmarks built on real-world MCP servers:

\begin{itemize}
    \item \textbf{MCPMark}~\citep{DBLP:journals/corr/abs-2509-24002} 
    (available at \url{https://mcpmark.ai/}) is a stress-testing 
    benchmark comprising 121 tasks across five MCP services with 
    curated environment instances. Each task pairs a 
    natural-language instruction with a programmatic verifier and 
    is executed from a task-specific starting snapshot. We adopt 
    its five services as follows:
    \begin{itemize}[leftmargin=*,nosep]
        \item \textbf{Filesystem}: file I/O, directory 
        organization, metadata inspection, and search over curated 
        folder structures mirroring everyday user scenarios 
        (\textit{e.g.}, desktop files, legal documents, student 
        databases).
        \item \textbf{GitHub}: project management and Git 
        operations including CI/CD, issues, branches, pull 
        requests, and commits, grounded in repositories with 
        realistic development histories.
        \item \textbf{Notion}: creating, editing, and querying 
        documents and databases via the official Notion API, 
        instantiated from widely adopted templates 
        (\textit{e.g.}, online resume, travel planner, student 
        dashboard).
        \item \textbf{Playwright}: browser automation including 
        navigation, form completion, data extraction, and 
        screenshot generation.
        \item \textbf{PostgreSQL}: schema exploration and SQL 
        query execution over representative template databases 
        with realistic schemas (\textit{e.g.}, Chinook, DVD 
        Rental, Employees, Lego).
    \end{itemize}

    \item \textbf{MCP-Universe}~\citep{DBLP:journals/corr/abs-2508-14704} 
    (available at \url{https://github.com/SalesforceAIResearch/MCP-Universe}) 
    is a benchmark of 231 tasks spanning six MCP service. We adopt three domains and 
    exclude three (3D Design, Repository Management, and Browser 
    Automation) that overlap with services already covered by 
    \textsc{MCPMark}. The three adopted MCP services are:
    \begin{itemize}[leftmargin=*,nosep]
        \item \textbf{Location Navigation}: geographic queries and 
        route planning via the Google Maps MCP server, including 
        distance optimization, stop finding, and multi-point 
        navigation.
        \item \textbf{Financial Analysis}: stock and financial 
        data analysis via the YFinance MCP server, covering 
        portfolio analysis, trading strategies, and institutional 
        holdings.
        \item \textbf{Web Search}: open-domain information 
        retrieval using Google Search and Fetch MCP servers, 
        spanning person identification, entity discovery, and 
        complex reasoning.
    \end{itemize}
\end{itemize}

\begin{table}[ht]
\centering
\small
\begin{tabular}{l|cccc}
\toprule
\rowcolor{gray!10}
\textbf{MCP Service} & \textbf{\#Tools} & \textbf{\#Instances} & \textbf{Train} & \textbf{Test} \\
\midrule
Filesystem   & 14 & 10 & 11 & 19 \\
GitHub       & 94 &  6 &  8 & 15 \\
Notion       & 19 & 10 & 10 & 18 \\
Playwright   & 21 &  3 &  6 & 13 \\
PostgreSQL   &  9 &  7 &  8 & 13 \\
\midrule
\textit{Total} & \textit{157} & \textit{36} & \textit{43} & \textit{78} \\
\bottomrule
\end{tabular}
\caption{Task distribution of \textsc{MCPMark} under the 
same-environment split.}
\label{tab:mcpmark_split}
\end{table}

\begin{table}[ht]
\centering
\small
\begin{tabular}{l|ccc}
\toprule
\rowcolor{gray!10}
\textbf{MCP Service} & \textbf{\#Tools} & \textbf{Train} & \textbf{Test} \\
\midrule
Location Navigation   & 7 &  9 & 26 \\
Financial Analysis    &                      10 & 10 & 30 \\
Web Search            &                      2 & 13 & 37 \\
\midrule
\textit{Total} & \textit{19} & \textit{32} & \textit{93} \\
\bottomrule
\end{tabular}
\caption{Task distribution of the adopted \textsc{MCP-Universe} 
domains under the same-environment split.}
\label{tab:mcpuniverse_split}
\end{table}

\begin{table*}[ht]
\centering
\small
\resizebox{\textwidth}{!}{
\begin{tabular}{l|cc|cc}
\toprule
\rowcolor{gray!10}
\textbf{MCP Service} & \textbf{Train Instances} & \textbf{\#Tasks} & \textbf{Test Instances} & \textbf{\#Tasks} \\
\midrule
Filesystem
  & file\_context, file\_property, & \multirow{2}{*}{13} & desktop, desktop\_template, & \multirow{2}{*}{17} \\
  & student\_database, threestudio & & folder\_structure, legal\_document, papers, votenet & \\
\midrule
GitHub
  & claude-code, easyr1, harmony & 14 & build\_your\_own\_x, mcpmark-cicd, missing-semester & 9 \\
\midrule
Notion
  & it\_trouble\_shooting\_hub, japan\_travel\_planner, & \multirow{2}{*}{11} & company\_in\_a\_box, cs\_student\_dashboard, & \multirow{2}{*}{17} \\
  & standard\_operating\_procedure, team\_projects & & online\_resume, python\_roadmap, self\_assessment, toronto\_guide & \\
\midrule
Playwright
  & reddit & 5 & shopping, shopping\_admin & 14 \\
\midrule
PostgreSQL
  & dvdrental, lego, security & 8 & chinook, employees, sports, vectors & 13 \\
\midrule
\textit{Total} & \textit{15 instances} & \textit{51} & \textit{21 instances} & \textit{70} \\
\bottomrule
\end{tabular}
}
\caption{\textsc{MCPMark} under the cross-environment split: entire 
environment instances are assigned to either train or test, so 
that no instance appears in both splits. The same 121 tasks as in 
Table~\ref{tab:mcpmark_split} are repartitioned along the 
instance axis. Train instances are used for brainstorm-based 
memory construction; test instances evaluate transfer to unseen 
environment instances.}
\label{tab:cross_env_split}
\end{table*}

\paragraph{Data Split Protocol.}\label{app:split}
We split each benchmark for two purposes, corresponding to two 
generalization settings:
\begin{itemize}[leftmargin=*,nosep]
    \item \textbf{Same-environment split.} 
    This is the most basic setting for memory accumulation: train 
    and test tasks \textit{share the same environment instances 
    and the same tool set}, isolating the effect of the memory 
    itself from any environment or tool shift. Within each MCP 
    service, tasks are randomly partitioned into disjoint train 
    and test sets at a ratio of roughly 1:2, while allowing both 
    splits to draw from the same pool of environment instances. 
    Each task still begins from its own task-specific snapshot, 
    so train and test tasks never overlap at the task level. 
    Statistics are summarized in Table~\ref{tab:mcpmark_split} 
    (\textsc{MCPMark}) and Table~\ref{tab:mcpuniverse_split} 
    (\textsc{MCP-Universe}).

    \item \textbf{Cross-environment split.} 
    This setting tests whether memory generalizes beyond the 
    environment instances seen during construction: train and test 
    tasks \textit{share the same tool set but no longer share 
    environment instances}. Within each MCP service, environment 
    instances themselves are randomly partitioned into disjoint 
    train and test groups at a ratio of roughly 2:3; all tasks 
    associated with a train-side instance are used for memory 
    construction, and all tasks associated with a test-side 
    instance are used for evaluation. The resulting task counts 
    therefore follow the instance assignment rather than a 
    pre-specified task-level ratio. \textsc{MCP-Universe} does not 
    define environment instances and is therefore excluded from 
    this protocol. The instance-level assignment is summarized 
    in Table~\ref{tab:cross_env_split}.
\end{itemize}

\paragraph{Metrics.} 
We report pass@1 and pass@4 as the primary 
evaluation metrics, following the standard protocol of 
\textsc{MCPMark}~\citep{DBLP:journals/corr/abs-2509-24002}. For 
each task, we run the agent four times independently from the 
same starting snapshot and pass the resulting trajectories 
through the task's programmatic verifier. Pass@1 
measures the average per-run success rate across the four runs, 
reflecting the agent's expected performance on a single attempt; 
Pass@4 measures whether \emph{at least one} of the four 
runs succeeds, reflecting the agent's coverage when multiple 
attempts are permitted. Higher is better for both metrics.

\subsection{Baseline Setup}\label{app:baseline}
In this section, we provide detailed descriptions of each baseline 
used in our comparison. 

\paragraph{Tool-side Optimization.}
These baselines optimize tool descriptions or generate usage 
examples in an offline training phase, then inject the resulting 
artifacts into the agent prompt at inference time.
\begin{itemize}[leftmargin=*,nosep]
    \item \textbf{EasyTool}~\citep{yuan2025easytool} 
    distills lengthy tool documentation into concise 
    \emph{tool instructions} through a two-stage pipeline: an LLM 
    first rewrites each tool's documentation into a unified 
    description of its general purpose and built-in functions, then 
    generates scenario and parameter-value demonstrations verified 
    by actually invoking the tool. We instantiate both stages with 
    the same backbone LLM used in the corresponding experiment, and 
    run the verification step on a clean environment snapshot to 
    avoid contaminating evaluation states.
    
    \item \textbf{Play2Prompt}~\citep{fang2025play2prompt} 
    optimizes tool documentation through a \emph{tool-play} 
    procedure that iteratively interacts with each tool and 
    reflects on its outputs. It runs two beam-search stages: the 
    first generates usage examples by exploring valid invocations 
    in a trial-and-error loop, and the second refines the tool 
    documentation using those examples as a validation set. We 
    instantiate both stages with the same backbone LLM used in the 
    corresponding experiment, and ground example generation in the 
    current MCP environment instance to ensure valid invocations 
    under its stateful resources.

    \item \textbf{ToolOptimal}\footnote{The original paper does not 
    name its method; we use ToolOptimal for brevity.}~\citep{wu2025joint} 
    jointly optimizes the agent's task instruction and each tool's 
    description through a three-stage verbalized loop: a feedback 
    generator evaluates the rollout, a suggestion coordinator emits 
    separate revisions for the instruction and the description, and 
    a context refiner aggregates a batch of revisions before 
    committing, analogous to gradient accumulation. We instantiate 
    all three stages with the same backbone LLM used in the 
    corresponding experiment, and feed the feedback generator both 
    the verifier outcome (effectiveness) and the tool-call trajectory 
    length and redundancy (efficiency) on the training split.
\end{itemize}

\paragraph{Experience-based Memory.}
These baselines accumulate experience from past task executions 
and inject it into the agent prompt for future tasks.
\begin{itemize}[leftmargin=*,nosep]
    \item \textbf{ExpeL}~\citep{zhao2024expel} 
    distills transferable \emph{insights} from training trajectories 
    through two channels: contrasting same-task success/failure pairs 
    and aggregating patterns across multiple successful trajectories; 
    each insight carries an importance count maintained by 
    add/upvote/downvote/edit operations and is removed when the count 
    reaches zero. Insight extraction is instantiated with the same 
    backbone LLM used in the corresponding experiment. At inference, 
    the full insight set is concatenated into the agent prompt, 
    together with the top-$k$ successful trajectories retrieved by 
    task-embedding similarity.

    \item \textbf{A-MEM}~\citep{xu2026mem} 
    follows a Zettelkasten-inspired design in which each new 
    memory triggers three operations: an LLM generates structured 
    attributes and embeddings (note construction); retrieval 
    identifies semantic neighbors and establishes bidirectional 
    links (link generation); and linked notes update their 
    representations to capture higher-order patterns (memory 
    evolution). A-MEM was originally designed for 
    long-conversation QA; to adapt it for MCP tool use, we 
    replace dialogue content with stateful task-execution 
    trajectories (task description, MCP tool-call sequence, and 
    success/failure outcome), and use the new task description as 
    the retrieval query.

    \item \textbf{ACE}~\citep{zhang2025agentic} 
    maintains an evolving \emph{playbook} represented as a collection 
    of structured, itemized bullets, each carrying content (a 
    reusable strategy or failure mode) and counters tracking how 
    often it was marked helpful or harmful. The playbook is updated 
    through a three-role loop: a Generator runs the task and flags 
    which bullets were useful or misleading, a Reflector distills 
    lessons from the rollout, and a Curator merges those lessons 
    into the playbook as incremental delta entries via deterministic 
    non-LLM logic, with a separate grow-and-refine step that 
    de-duplicates bullets by semantic similarity. At inference, the 
    full playbook is provided to the Generator as context, rather 
    than retrieving top-$k$ entries. The Generator, Reflector, and 
    Curator are all instantiated with the same backbone LLM used in 
    the corresponding experiment. Since MCP verifiers provide 
    deterministic pass/fail signals, we use them to drive the 
    helpful/harmful counter bumps on the training split.
\end{itemize}

\subsection{Agent Framework Setup}\label{app:agent-setup}
In this section, we detail the setups of our three adopted agent 
frameworks, ReAct, CodeAct, and SwitchAct:

\subsubsection{ReAct}
ReAct~\citep{yao2022react} interleaves natural-language 
reasoning with structured tool invocations, alternating between 
\emph{thought} steps that plan the next action and \emph{action} 
steps that issue a tool call. Its action space is restricted to 
single-tool invocations.

\subsubsection{CodeAct}
CodeAct~\citep{wang2024executable} unifies the action space into 
executable Python code: at each step the agent emits a Python 
snippet that may compose multiple tools, perform control flow, 
and post-process intermediate results, and observes the runtime 
output rather than the return value of a single function call.

\subsubsection{SwitchAct}
SwitchAct is a simple hybrid agent we implement to serve as the 
default backbone in our experiments: by combining both action 
interfaces above, it grants the agent comprehensive tool-use 
ability and allows memory to be accumulated from a broader 
spectrum of behaviors than either single-mode agent alone. 
Concretely, SwitchAct exposes \emph{both} action interfaces to 
the LLM in parallel, letting the model autonomously choose, at 
each step, between a single-tool function call (as in ReAct) 
and a Python snippet (as in CodeAct). SwitchAct serves as the 
default agent backbone for our main results 
(\S\ref{sec:exp:rq1}); ReAct and CodeAct are used as target 
agents for the cross-agent generalization experiments in 
\S\ref{sec:exp:rq2}.

\subsection{LLM Backbones}\label{app:llm-backbone}
We evaluate  with three proprietary LLM backbones 
spanning two model families: \textbf{GPT-5.4} 
(\texttt{gpt-5.4-20260305}), \textbf{GPT-5.4-mini} 
(\texttt{gpt-5.4-mini-20260317}), and \textbf{Grok-4} 
(\texttt{grok-4-2025-04-01-preview}). All three models are 
accessed through their respective official APIs. For the two 
GPT-5.4 backbones we set the reasoning effort to \texttt{medium} 
to control for inference-time compute; Grok-4 does not expose a 
reasoning-effort knob, and we use its default reasoning setting.

\section{Implementation Details}
\label{app:hyperparams}

For Memory Bootstrapping, we generate $k{=}3$ seed tasks per tool and 
run $N{=}4$ rollouts per task. Capability Exploration runs for $R{=}3$ 
rounds; in each round, the proposer generates $3$ boundary tasks and 
$3$ affordance tasks for every target tool. At inference, 
Dynamic Memory Traversal retrieves $k_r{=}3$ seed trace nodes and 
operates under a read budget of $B{=}8$ steps.

\section{Additional Experiment Result}
\subsection{\textsc{MCPMark} results using Grok-4}
Table~\ref{tab:rq1_mcpmark_grok} presents the results on \textsc{MCPMark} using the Grok-4 as the LLM backbone. \mname{} consistently outperforms the strongest baseline. 

\begin{table*}[!ht]
\centering
\fontsize{7.5pt}{8pt}\selectfont
\setlength{\tabcolsep}{2.2pt}
\renewcommand{\arraystretch}{1.25}
\begin{tabular}{l|cc|cc|cc|cc|cc|cc}
\toprule
\multirow{2}{*}{\textbf{Method}}
& \multicolumn{2}{c|}{\textbf{Filesystem}}
& \multicolumn{2}{c|}{\textbf{GitHub}}
& \multicolumn{2}{c|}{\textbf{Notion}}
& \multicolumn{2}{c|}{\textbf{Playwright}}
& \multicolumn{2}{c|}{\textbf{PostgreSQL}}
& \multicolumn{2}{c}{\textbf{Overall}} \\
& pass@1$\uparrow$ & pass@4$\uparrow$
& pass@1$\uparrow$ & pass@4$\uparrow$
& pass@1$\uparrow$ & pass@4$\uparrow$
& pass@1$\uparrow$ & pass@4$\uparrow$
& pass@1$\uparrow$ & pass@4$\uparrow$
& pass@1$\uparrow$ & pass@4$\uparrow$ \\
\midrule
\rowcolor{gray!10}
\multicolumn{13}{c}{\textit{Grok-4}} \\
\midrule
Vanilla
& 55.26 & 73.68 & 11.67 & 33.33 & 4.17 & 11.11 & 11.54 & 30.77 & 48.08 & 61.54 & 26.14 & 42.09 \\
EasyTool
& 63.16 & \hlsecond{89.47} & 18.33 & \hlsecond{40.00} & \hlsecond{8.33} & \hlsecond{22.22} & 9.62 & 30.77 & 48.08 & \hlsecond{69.23} & 29.50 & \hlsecond{50.34} \\
Play2Prompt
& 56.58 & 78.95 & 15.00 & 33.33 & \hlsecond{8.33} & 16.67 & 13.46 & 23.08 & \hlsecond{50.00} & \hlsecond{69.23} & 28.67 & 44.25 \\
ToolOptimal
& 59.21 & 84.21 & 15.00 & \hlsecond{40.00} & 6.94 & \hlsecond{22.22} & 7.69 & 23.08 & 46.15 & 53.85 & 27.00 & 44.67 \\
Expel
& 55.26 & 68.42
& 13.33 & 33.33
& 5.56 & 16.67
& 25.00 & 30.77
& \hlsecond{50.00} & 61.54
& 29.83 & 42.15 \\
A-Mem
& 55.26 & 78.95
& \hlsecond{23.33} & \hlsecond{40.00}
& 5.56 & \hlsecond{22.22}
& 15.38 & 23.08
& 26.92 & 61.54
& 25.29 & 45.16 \\
ACE
& \hlsecond{64.47} & 84.21
& 16.67 & 33.33
& 6.94 & \hlsecond{22.22}
& \hlsecond{26.92} & \hlsecond{38.46}
& 48.08 & 53.85
& \hlsecond{32.62} & 46.41 \\
\rowcolor[HTML]{F0F6FF}
\mname{}
& \hlfirst{73.68} & \hlfirst{94.74} & \hlfirst{26.67} & \hlfirst{46.67} & \hlfirst{15.28} & \hlfirst{27.78} & \hlfirst{28.85} & \hlfirst{46.15} & \hlfirst{53.85} & \hlfirst{76.92} & \hlfirst{39.67} & \hlfirst{58.45} \\
\textit{Impro(\%)}
& \textit{+14.29} & \textit{+5.89} & \textit{+14.32} & \textit{+16.68} & \textit{+83.43} & \textit{+25.02} & \textit{+7.17} & \textit{+19.99} & \textit{+7.70} & \textit{+11.11} & \textit{+21.61} & \textit{+16.11} \\
\bottomrule
\end{tabular}
\captionsetup{font=footnotesize}
\caption{Performance comparison on \textsc{MCPMark} under the same-environment split, using \textit{SwitchAct} with Grok-4 as the backbone. The \textbf{best} and \underline{second best} results are marked in bold and with underline, respectively. \textit{Impro(\%)} is the relative gain of \mname{} over the strongest baseline.}
\label{tab:rq1_mcpmark_grok}
\vspace{-0.4cm}
\end{table*}
\subsection{\textsc{MCP-Universe} results}

\begin{table*}[!ht]
\centering
\fontsize{7.5pt}{8pt}\selectfont
\setlength{\tabcolsep}{2.4pt}
\renewcommand{\arraystretch}{1.25}
\begin{tabular}{l|cc|cc|cc|cc}
\toprule
\multirow{2}{*}{\textbf{Method}}
& \multicolumn{2}{c|}{\textbf{Loc.\ Navigation}}
& \multicolumn{2}{c|}{\textbf{Financial Analysis}}
& \multicolumn{2}{c|}{\textbf{Web Searching}}
& \multicolumn{2}{c}{\textbf{Overall}} \\
& pass@1$\uparrow$ & pass@4$\uparrow$
& pass@1$\uparrow$ & pass@4$\uparrow$
& pass@1$\uparrow$ & pass@4$\uparrow$
& pass@1$\uparrow$ & pass@4$\uparrow$ \\
\midrule
Vanilla
& 30.77 & 46.15
& 77.50 & 83.33
& 41.22 & 59.46
& 49.83 & 62.98 \\
EasyTool
& 35.58 & 53.85
& 73.33 & 83.33
& \hlsecond{47.30} & 59.46
& 52.07 & 65.55 \\
Play2Prompt
& 32.69 & 50.00
& 71.67 & 86.67
& 45.95 & 62.16
& 50.10 & 66.28 \\
ToolOptimal
& 31.73 & 50.00
& 75.00 & 86.67
& 46.62 & 62.16
& 51.12 & 66.28 \\
Expel
& 39.42 & 46.15
& 78.33 & 86.67
& 42.57 & \hlsecond{64.86}
& 53.44 & 65.89 \\
A-Mem
& 37.50 & 53.85
& 71.67 & 83.33
& 45.95 & 59.46
& 51.71 & 65.55 \\
ACE
& \hlsecond{42.31} & \hlsecond{57.69}
& \hlsecond{79.17} & \hlsecond{90.00}
& 43.24 & 62.16
& \hlsecond{54.91} & \hlsecond{69.95} \\
\rowcolor[HTML]{F0F6FF}
\mname{}
& \hlfirst{47.12} & \hlfirst{65.38}
& \hlfirst{82.50} & \hlfirst{93.33}
& \hlfirst{52.03} & \hlfirst{70.27}
& \hlfirst{60.55} & \hlfirst{76.33} \\
\textit{Impro(\%)}
& \textit{+11.37} & \textit{+13.33}
& \textit{+4.21} & \textit{+3.70}
& \textit{+10.00} & \textit{+8.34}
& \textit{+10.27} & \textit{+9.12} \\
\bottomrule
\end{tabular}
\vspace{-0.2cm}
\captionsetup{font=footnotesize}
\caption{Performance comparison on \textsc{MCP-Universe} under the same-environment split, using \textit{SwitchAct} with GPT-5.4 as the backbone. The \textbf{best} and \underline{second best} results are marked in bold and with underline, respectively. \textit{Impro(\%)} is the relative gain of \mname{} over the strongest baseline.}
\label{tab:rq1_mcpuniverse}
\end{table*}
Table~\ref{tab:rq1_mcpuniverse} presents the results on \textsc{MCP-Universe} using GPT-5.4 as the LLM backbone. \mname{} consistently outperforms the strongest baseline. 

\subsection{Frontier Exploration Effectiveness}
\label{app:ablation_frontier}

\begin{figure}[t]
  \centering
  \includegraphics[width=0.48\textwidth]{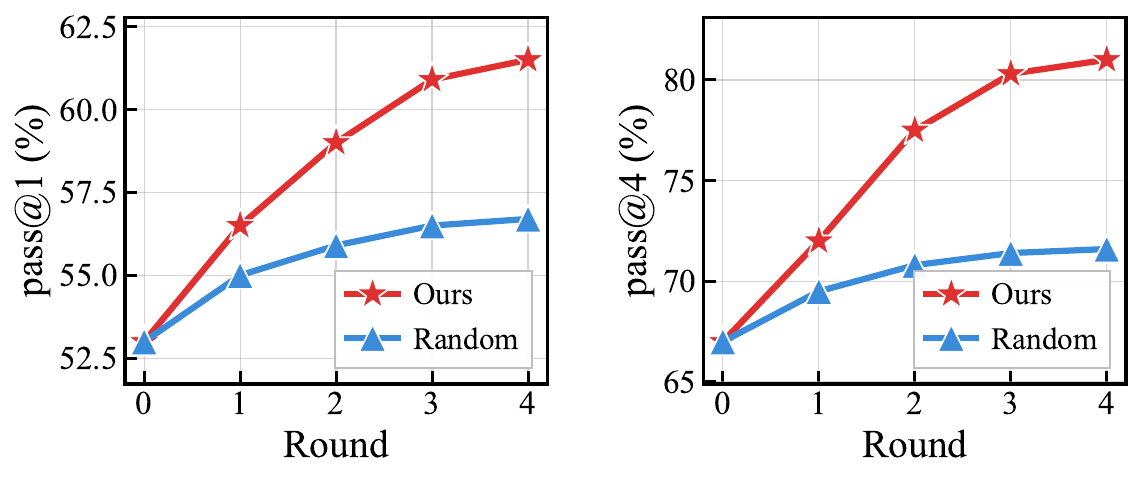}
  \vspace{-1.9em}
  \captionsetup{font=footnotesize}
  \caption{Effect of frontier exploration across brainstorm
  rounds on \textsc{MCPMark}.}
  \label{fig:rq3_rounds}
\end{figure}

\Cref{fig:rq3_rounds} compares the Frontier Exploration of \mname{} with a baseline exploration approach that uses the same per-round exploration budget but only samples from seed tasks.

Frontier exploration avoids redundant
memory construction. Both methods improve as the number of brainstorm rounds $R$
increases, but their trajectories differ sharply. Random improves
early and then plateaus, while \mname{} continues to improve
across rounds, widening the gap as $R$ grows. This pattern
reveals a failure mode we call \emph{frontier blindness}: without
an explicit estimate of what is already known about each tool,
the LLM tends to propose overlapping tasks that revisit familiar
tool behaviors. As a result, additional random exploration
generates redundant experience. In contrast, \mname{} directs
later rounds toward underexplored capabilities and boundary
conditions, allowing each round to contribute new
execution-verified knowledge. We set $R=3$ by default, which
captures most of the observed gain while keeping exploration cost
moderate.

\section{Computational Resources and Software Environment}
All experiments were executed on a CPU-only server featuring four Intel Xeon E7-4830 v3 processors (48 physical cores, 96 threads at 2.10\,GHz base frequency) with 503\,GB of main memory, running Ubuntu 24.04.1 LTS (kernel 6.8.0). Since \mname{} relies entirely on cloud-hosted LLM APIs for both brainstorm construction and task execution, no local GPU resources were required.  The software environment was managed through Conda 25.9.1 with Python 3.12. Core API communication was handled by the OpenAI Python SDK (v1.109.1) and the MCP protocol library (v1.27.0). Benchmark evaluation on \textsc{MCPMark} (v0.0.1) leveraged several environment-specific connectors, including Playwright (v1.58.0) for browser automation, \texttt{psycopg2} (v2.9.11) for PostgreSQL interaction, and \texttt{notion-client} (v2.4.0) for Notion API access. Token counting was performed with \texttt{tiktoken} (v0.12.0), and asynchronous HTTP requests were dispatched via \texttt{httpx} (v0.28.1). Data processing relied on NumPy (v2.4.4) and standard scientific Python utilities.

\section{The Use of Large Language Models}
During the preparation of this manuscript, Large Language Models (LLMs) were employed to assist with language polishing, grammatical refinement, and coding support (e.g., debugging scripts and setting up experimental environments). Every piece of LLM-generated content was manually reviewed and validated by the authors prior to inclusion. The core research design, experimental analysis, and all scientific conclusions were carried out independently by the authors without LLM involvement.

\clearpage
\onecolumn

\section{Case Study}
\label{app:case-seed-round}

\tcbset{
  casebox/.style={
    breakable,
    coltitle=black,
    fonttitle=\bfseries,
    boxrule=0.45pt,
    left=1.5mm,
    right=1.5mm,
    top=1mm,
    bottom=1mm,
    boxsep=1mm,
    arc=1mm,
    before skip=4pt,
    after skip=4pt
  },
  seedbox/.style={
    casebox,
    colback=blue!2!white,
    colframe=blue!45!black,
    colbacktitle=blue!12!white
  },
  tracebox/.style={
    casebox,
    colback=violet!2!white,
    colframe=violet!45!black,
    colbacktitle=violet!12!white
  },
  toolbox/.style={
    casebox,
    colback=green!2!white,
    colframe=green!45!black,
    colbacktitle=green!12!white
  },
  stratbox/.style={
    casebox,
    colback=teal!2!white,
    colframe=teal!45!black,
    colbacktitle=teal!12!white
  },
  frontierbox/.style={
    casebox,
    colback=orange!3!white,
    colframe=orange!55!black,
    colbacktitle=orange!14!white
  },
  navbox/.style={
    casebox,
    colback=purple!2!white,
    colframe=purple!45!black,
    colbacktitle=purple!12!white
  },
  effectbox/.style={
    casebox,
    colback=gray!3!white,
    colframe=gray!45!black,
    colbacktitle=gray!15!white
  }
}

\begin{tcolorbox}[seedbox,title={Memory Bootstrapping}]
\begin{itemize}[leftmargin=*,nosep]
    \item \textbf{Target tool:} \texttt{list\_directory\_with\_sizes}
    \item \textbf{Web search scenario:} ranking sibling folders by storage footprint
    \item \textbf{Generated task:} find the student folder(s) with the smallest and largest total stored bytes; write totals and names to \texttt{student\_directory\_size\_extremes.txt}
\end{itemize}
\end{tcolorbox}

\begin{tcolorbox}[tracebox,title={Memory Induction for \(\mathcal{G}_\mathsf{trace}\)}]
\textbf{Task.} Find the smallest and largest student folder by total bytes; write the result to \texttt{student\_directory\_size\_extremes.txt}.

\medskip
\textbf{4 rollouts.}
\begin{itemize}[leftmargin=*,nosep]
    \item Attempt 1 (\(\times\)): 28 calls --- looped between \texttt{list\_directory} / \texttt{search\_files} / \texttt{get\_file\_info}, took shallow sizes at face value.
    \item Attempt 2 (\(\times\)): 36 calls --- repeated \texttt{list\_directory\_with\_sizes} four times on the same parent without summing recursively.
    \item Attempt 3 (\(\times\)): 175+ calls --- stuck in a \texttt{list\_directory\_with\_sizes} loop, eventually wrote a wrong report.
    \item Attempt 4 (\(\checkmark\)): 24 calls --- \texttt{list\_allowed\_dirs} \(\rightarrow\) \texttt{directory\_tree} \(\rightarrow\) \texttt{get\_file\_info} \(\rightarrow\) \texttt{search\_files} \(\rightarrow\) \texttt{list\_directory\_with\_sizes} \(\rightarrow\) \texttt{get\_file\_info} \(\rightarrow\) \texttt{write\_file} \(\rightarrow\) \texttt{read\_text\_file}.
\end{itemize}

\medskip
\textbf{1. Agent-neutral trace \(\tilde\xi_i\) (per-task compression of the 4 rollouts)}
\begin{enumerate}[leftmargin=*,nosep]
    \item \texttt{list\_allowed\_directories} --- confirm the writable workspace before any inspection
    \item \texttt{directory\_tree} --- survey top-level structure to enumerate candidate folders
    \item \texttt{get\_file\_info} --- validate target metadata before computing
    \item \texttt{search\_files} --- recursive discovery so no candidate folder is missed
    \item \texttt{list\_directory\_with\_sizes} --- obtain per-subdirectory totals for comparison
    \item \texttt{get\_file\_info} --- re-check exact byte totals on the identified extremes
    \item \texttt{write\_file} --- write the summary file
    \item \texttt{read\_text\_file} --- verify the saved note matches the intended output
\end{enumerate}
\end{tcolorbox}

\begin{tcolorbox}[toolbox,title={Memory Induction for \(\mathcal{G}_\mathsf{cap}\)}]
\textbf{Input.} All query nodes whose refined trace touches \texttt{list\_directory\_with\_sizes}. For each such query we feed in the refined trace plus any \texttt{task\_tips} mentioning the tool:

\medskip
\textbf{Output} (new fields appended to \texttt{tools["list\_directory\_with\_sizes"]}):

\medskip
\begin{flushleft}
\ttfamily\small
Input:\par
\medskip
[q\_student\_directory\_size\_extremes\_note] summary: rank sibling folders by total bytes...\par
\hspace*{1em}Refined trace:\par
\hspace*{2em}1. list\_allowed\_directories --- confirm writable workspace\par
\hspace*{2em}2. directory\_tree \hspace*{4.3em}--- enumerate candidate folders\par
\hspace*{2em}3. get\_file\_info \hspace*{4.1em}--- validate target metadata\par
\hspace*{2em}4. search\_files \hspace*{4.9em}--- recursive discovery\par
\hspace*{2em}5. list\_directory\_with\_sizes --- obtain per-subdirectory totals\par
\hspace*{2em}6. get\_file\_info \hspace*{4.1em}--- re-check exact byte totals on extremes\par
\hspace*{2em}7. write\_file \hspace*{5.7em}--- write summary\par
\hspace*{2em}8. read\_text\_file \hspace*{3.7em}--- verify saved output\par
\hspace*{1em}Related tips: shallow per-entry sizes are not recursive totals;\par
\hspace*{5.4em}verify extremes with get\_file\_info before writing.\par
[q\_\ldots{} ] \ldots\par
\medskip
Output:\par
\medskip
\{\par
\hspace*{1em}"new\_affordance": [\par
\hspace*{2em}\{"text": "Sum child entries explicitly when ranking siblings --- the reported sizes are shallow.",\par
\hspace*{2.4em}"source\_queries": ["q\_student\_directory\_size\_extremes\_note", \ldots]\}\par
\hspace*{1em}],\par
\hspace*{1em}"new\_boundaries": [\par
\hspace*{2em}\{"text": "Per-entry sizes are non-recursive; do not treat them as folder totals.",\par
\hspace*{2.4em}"source\_queries": ["q\_student\_directory\_size\_extremes\_note"]\}\par
\hspace*{1em}],\par
\hspace*{1em}"new\_co\_usage": [\par
\hspace*{2em}\{"text": "list\_directory\_with\_sizes \(\rightarrow\) get\_file\_info to confirm exact byte totals on extremes.",\par
\hspace*{2.4em}"related\_tools": ["list\_directory\_with\_sizes", "get\_file\_info"],\par
\hspace*{2.4em}"source\_queries": ["q\_student\_directory\_size\_extremes\_note"]\}\par
\hspace*{1em}]\par
\}\par
\end{flushleft}
\end{tcolorbox}

\begin{tcolorbox}[stratbox,title={Memory Induction for \(\mathcal{G}_\mathsf{strat}\)}]
\textbf{Input.} A subgraph of semantically related query nodes (here: workspace-rooted file-system reporting tasks). For each query we feed in the refined trace plus its \texttt{task\_tips}:

\medskip
\textbf{Output} (new entry appended to \texttt{strategy}):

\medskip
\begin{flushleft}
\ttfamily\small
Input:\par
\medskip
[q\_student\_directory\_size\_extremes\_note] summary: rank sibling folders by total bytes\par
\hspace*{1em}Refined trace: list\_allowed\_directories \(\rightarrow\) directory\_tree \(\rightarrow\) get\_file\_info \(\rightarrow\)\par
\hspace*{8.6em}search\_files \(\rightarrow\) list\_directory\_with\_sizes \(\rightarrow\) get\_file\_info \(\rightarrow\)\par
\hspace*{8.6em}write\_file \(\rightarrow\) read\_text\_file\par
\hspace*{1em}Task tips: start by confirming the writable workspace; only then explore.\par
[q\_largest\_log\_file\_per\_service] \hspace*{1em}Refined trace: list\_allowed\_directories \(\rightarrow\) directory\_tree \(\rightarrow\) \ldots{} ; tips: \ldots\par
[q\_count\_python\_files\_under\_assets] Refined trace: list\_allowed\_directories \(\rightarrow\) search\_files \(\rightarrow\) \ldots{} ; tips: \ldots\par
[q\_\ldots{} ] \ldots\par
\medskip
Output:\par
\medskip
\{\par
\hspace*{1em}"new\_strategy": [\par
\hspace*{2em}\{"text": "Begin with workspace and constraint discovery before acting: identify allowed roots and writable locations first so later steps target the correct scope.",\par
\hspace*{2.4em}"source\_queries": ["q\_student\_directory\_size\_extremes\_note",\par
\hspace*{10.3em}"q\_largest\_log\_file\_per\_service",\par
\hspace*{10.3em}"q\_count\_python\_files\_under\_assets", \ldots]\}\par
\hspace*{1em}]\par
\}\par
\end{flushleft}
\end{tcolorbox}

\begin{tcolorbox}[frontierbox,title={Capability Exploration}]
\begin{flushleft}
\ttfamily\small
Sampled knowledge-graph neighborhood (centered on list\_directory\_with\_sizes):\par
\medskip
nodes:\par
\hspace*{1em}n1 = Tool\{name: list\_directory\_with\_sizes, role: center\}\par
\hspace*{1em}n2 = Tool\{name: get\_file\_info, role: verifier\}\par
\hspace*{1em}n3 = Tool\{name: directory\_tree, role: structure\_survey\}\par
\hspace*{1em}n4 = Tool\{name: search\_files, role: recursive\_discovery\}\par
\medskip
typed edges:\par
\hspace*{1em}(n1) -[co\_usage: confirm exact byte totals]-\(\rightarrow\) (n2)\par
\hspace*{1em}(n3) -[co\_usage: enumerate candidate folders]-\(\rightarrow\) (n1)\par
\hspace*{1em}(n4) -[co\_usage: discover recursive candidates]-\(\rightarrow\) (n3)\par
\hspace*{1em}(n4) -[co\_usage: verify discovered paths]-\(\rightarrow\) (n2)\par
\medskip
graph triples shown to proposer:\par
\hspace*{1em}\(\langle\)list\_directory\_with\_sizes, co\_usage, get\_file\_info\(\rangle\)\par
\hspace*{1em}\(\langle\)directory\_tree, co\_usage, list\_directory\_with\_sizes\(\rangle\)\par
\hspace*{1em}\(\langle\)search\_files, co\_usage, directory\_tree\(\rangle\)\par
\hspace*{1em}\(\langle\)search\_files, co\_usage, get\_file\_info\(\rangle\)\par
\medskip
Extracted knowledge from the subgraph nodes (the knowledge\_map actually shown to the proposer LLM):\par
\medskip
\#\# list\_directory\_with\_sizes\par
\hspace*{1em}affordance:\par
\hspace*{2em}- Sum child entries explicitly when ranking siblings --- sizes are shallow.\par
\hspace*{1em}boundaries:\par
\hspace*{2em}- Per-entry sizes are non-recursive; do not treat them as folder totals.\par
\hspace*{1em}co\_usage:\par
\hspace*{2em}- list\_directory\_with\_sizes \(\rightarrow\) get\_file\_info to confirm exact byte totals on extremes.\par
\medskip
\#\# get\_file\_info\par
\hspace*{1em}affordance:\par
\hspace*{2em}- Authoritative byte counts on a single path.\par
\hspace*{2em}- \ldots\par
\hspace*{1em}boundaries: \ldots\par
\hspace*{1em}co\_usage: \hspace*{1em}\ldots\par
\medskip
\#\# directory\_tree\par
\hspace*{1em}affordance: \ldots\par
\hspace*{1em}boundaries: \ldots\par
\hspace*{1em}co\_usage:\par
\hspace*{2em}- Pair with list\_directory\_with\_sizes / get\_file\_info for recursive discovery.\par
\hspace*{2em}- \ldots\par
\medskip
\#\# search\_files\par
\hspace*{1em}affordance: \ldots\par
\hspace*{1em}boundaries: \ldots\par
\hspace*{1em}co\_usage: \hspace*{1em}\ldots\par
\end{flushleft}
\medskip
\textbf{Two proposed tasks} (same target tool, opposite directions):
\begin{itemize}[leftmargin=*,nosep]
    \item \textbf{R2 --- boundary}: On a workspace with 10k+ deeply-nested subfolders and symlink cycles, rank the top-5 folders by \emph{recursive} byte total using \texttt{list\_directory\_with\_sizes}; write the ranking to \texttt{deep\_tree\_rank.txt}. \emph{(Goal: push the shallow-size limitation until it breaks.)}
    \item \textbf{R3 --- affordance}: Using \texttt{list\_directory\_with\_sizes} together with \texttt{get\_file\_info}, identify the single largest \emph{file} (not folder) under each top-level student directory and write \texttt{largest\_file\_per\_student.txt}. \emph{(Goal: exercise an untested file-level use combined with the known co-usage partner.)}
\end{itemize}
\end{tcolorbox}

\begin{tcolorbox}[navbox,title={Dynamic Memory Traversal}]
\textbf{New task (held-out).} \emph{For every top-level student folder, find the file with the most recent modification time and write \texttt{path<TAB>mtime} lines to \texttt{latest\_file\_per\_student.txt}.}

\medskip
\textbf{Step 1 --- embedding match to seeds}

Encode the task summary, retrieve top-3 query nodes by cosine similarity:

\medskip
\textbf{Step 2 --- navigator tool calls}
\begin{enumerate}[leftmargin=*,nosep]
    \item \texttt{ReadTrace} with argument \texttt{q\_student\_directory\_size\_extremes\_note} --- refined trace + task\_tips
    \item \texttt{Expand} with argument \texttt{q\_student\_directory\_size\_extremes\_note} --- similar query \texttt{q\_largest\_log\_file\_per\_service}
    \item \texttt{ReadTrace} with argument \texttt{q\_largest\_log\_file\_per\_service} --- trace exhibiting per-group extremum pattern
    \item \texttt{ReadTool} with argument \texttt{get\_file\_info} --- affordance entries (authoritative byte/mtime on a single path)
    \item \texttt{ReadStrategy} with argument \texttt{q\_student\_directory\_size\_extremes\_note} --- strategy \(i_{001}\) (workspace discovery first)
\end{enumerate}

\medskip
\textbf{Step 3 --- emitted guidance (returned to executor)}

\medskip
\begin{flushleft}
\ttfamily\small
SEED CANDIDATES (top-3 by embedding similarity):\par
\hspace*{1em}- q\_student\_directory\_size\_extremes\_note \hspace*{1em}(sim=0.81): rank sibling folders by total bytes\par
\hspace*{1em}- q\_largest\_log\_file\_per\_service \hspace*{2.7em}(sim=0.73): per-group extremum over a file attribute\par
\hspace*{1em}- q\_count\_python\_files\_under\_assets \hspace*{2.3em}(sim=0.61): recursive per-folder aggregation\par
\medskip
Past-task playbook:\par
\hspace*{1em}1. list\_allowed\_directories --- confirm writable workspace\par
\hspace*{1em}2. directory\_tree \hspace*{4.3em}--- enumerate top-level student folders\par
\hspace*{1em}3. search\_files \hspace*{4.9em}--- recursive file discovery under each folder\par
\hspace*{1em}4. get\_file\_info \hspace*{4.1em}--- read mtime per file (authoritative attribute)\par
\hspace*{1em}5. write\_file \hspace*{5.7em}--- write path<TAB>mtime lines\par
\hspace*{1em}6. read\_text\_file \hspace*{3.7em}--- verify saved output\par
\medskip
Strategy: begin with workspace and constraint discovery before acting (i\_001).\par
Tool tips:\par
\hspace*{1em}- get\_file\_info \(\rightarrow\) authoritative byte/mtime on a single path; use to resolve per-group extrema.\par
\hspace*{1em}- list\_directory\_with\_sizes \(\rightarrow\) not needed here (per-entry sizes non-recursive, and task is file-level).\par
\end{flushleft}
\end{tcolorbox}

\begin{tcolorbox}[effectbox,title={Step 4 --- execution}]
Executor follows the guidance and finishes in a single pass; verifier accepts byte-exact.

\begin{enumerate}[leftmargin=*,nosep]
    \item \texttt{list\_allowed\_directories} --- Strategy \(i_{001}\) --- workspace discovery first
    \item \texttt{directory\_tree(/workspace/students)} --- Playbook step 2 --- enumerate top-level student folders
    \item \texttt{search\_files(student\_01)} --- Playbook step 3 --- recursive file discovery
    \item \texttt{get\_file\_info(...)} \(\times N\) --- Tool tip --- \texttt{get\_file\_info} is the authoritative source for mtime
    \item (repeat 3--4 for each student folder) --- per-group extremum pattern carried over from \texttt{q\_largest\_log\_file\_per\_service}
    \item \texttt{write\_file(latest\_file\_per\_student.txt)} --- Playbook step 5
    \item \texttt{read\_text\_file(latest\_file\_per\_student.txt)} --- Playbook step 6 --- verify saved output
\end{enumerate}

\medskip
\textbf{Effect of guidance.} \texttt{list\_directory\_with\_sizes} --- the tool that broke 3/4 rollouts in the seed round --- never appears, because the Tool-tips line explicitly excluded it for this file-level variant. The executor also skips the shallow size-summing loop that consumed 175+ calls in seed-round attempt 3.
\end{tcolorbox}

\clearpage
\twocolumn

\newpage
\section{Prompt Set}
\label{sec: prompts} 
This section lists the core prompts used by \mname{}, 
grouped by the methodological stage they support: 
seed task proposal in Memory Bootstrapping, 
probe task proposal in Capability Exploration, 
trace reflection and capability/strategy aggregation in Memory Induction, 
and graph navigation in Dynamic Memory Traversal.

\begin{tcolorbox}[
  colback=lightgray!20,
  colframe=darkgray!80,
  title=Memory Bootstrapping Prompt,
  breakable,
  left=2mm, right=2mm,
  top=1mm, bottom=1mm]
\label{sec:prompt:seed}
You are a task designer for a benchmark that evaluates 
LLM agents on MCP tool usage.
\textbf{Goal:} Propose ONE new seed task that is feasible, 
non-trivial, covers a representative use of the target tool 
\texttt{\{target\_tool.name\}}, and is best solved by it.

\textbf{Input:}
\begin{itemize}[nosep,leftmargin=*]
  \item Target tool: \texttt{\{target\_tool.name\}}, 
    with description \{target\_tool.description\} and 
    input schema \{target\_tool.schema\}
  \item Real-world usage snippets retrieved via web 
    search for grounding: \{web\_search\_context\}
  \item One few-shot example for format reference (do 
    NOT copy): \{few\_shot\_task\}
\end{itemize}
\textbf{Output (JSON):}
\begin{itemize}[nosep,leftmargin=*]
  \item \texttt{task\_id}: snake\_case identifier
  \item \texttt{summary}: one-sentence summary
  \item \texttt{tool\_used}: 
    \texttt{\{target\_tool.name\}}
  \item \texttt{description\_md}: Markdown task 
    description with Title, Task Description, 
    Objectives, Constraints, and Expected Output
  \item \texttt{verify\_py}: self-contained Python 
    verifier that exits 0 on success or 1 on failure
\end{itemize}
Return ONLY the JSON object.
\end{tcolorbox}

\begin{tcolorbox}[
  colback=lightgray!20,
  colframe=darkgray!80,
  title=Tool-Trace Graph Induction Prompt,
  breakable,
  left=2mm, right=2mm,
  top=1mm, bottom=1mm]
\label{sec:prompt:reflection}
You are analyzing execution traces from a
tool-using agent to extract reusable knowledge.

\textbf{Goal:} Convert raw rollouts of a task into
an \texttt{agent\_neutral\_trace} and
\texttt{task\_level\_tips}, which together form a
trace node $q_{ij} = (Q_{ij}, \tilde{\xi}_{ij}, \mathcal{P}_{ij})$.

\textbf{Input:}
\begin{itemize}[nosep,leftmargin=*]
  \item Task description: \{task\_summary\}
  \item Category: \{category\}
  \item Rollouts from multiple attempts:
    \{rollouts\_text\}
  \item Real MCP tool namespace and allowed tool
    names: \{tool\_namespace\_hint\}
\end{itemize}
\textbf{Abstraction rule:} Produce
environment-invariant and agent-neutral knowledge.
Do NOT mention specific filenames, page IDs, table
names, issue numbers, schema fields, literal
values, paths, or attempt indices, and do NOT
preserve agent-specific action syntax, retry
patterns, or reasoning style.
\textbf{Output (JSON):}
\begin{itemize}[nosep,leftmargin=*]
  \item \texttt{agent\_neutral\_trace}: an ordered
    sequence of \texttt{(tool, rationale)} pairs
    synthesized across all rollouts, using the most
    successful attempt as the backbone when
    available and incorporating useful corrections
    from failed or partial attempts. \texttt{tool}
    is a real MCP tool name; \texttt{rationale}
    captures the tool-use intent of the step.
  \item \texttt{task\_level\_tips}: reusable
    task-level learnings distilled from successful
    and failed rollouts, not tied to one step.
    Rewrite failure lessons as positive cautions.
\end{itemize}
\noindent\texttt{\{} \\
\quad \texttt{"agent\_neutral\_trace": [\{"tool": "<real\_mcp\_tool>",} \\
\quad\quad \texttt{"rationale": "<tool-use intent of this step>"\}],} \\
\quad \texttt{"task\_level\_tips": ["<env-invariant tip>"]} \\
\texttt{\}}
Return ONLY the JSON object.
\end{tcolorbox}

\begin{tcolorbox}[
  colback=lightgray!20,
  colframe=darkgray!80,
  title=Tool-Capability Graph Induction Prompt,
  breakable,
  left=2mm, right=2mm,
  top=1mm, bottom=1mm]
\label{sec:prompt:tool-capability-induction}
You are analyzing how tool \texttt{\{tool\_name\}}
has been used across multiple tasks.

\textbf{Goal:} Update the Tool-Capability Graph
$\mathcal{G}_{\text{cap}}$ by inducing new
affordance entries $\mathcal{U}_i$, boundary
entries $\mathcal{B}_i$, and co-usage entries
$\mathcal{C}_i$ for tool node $t_i$ from usage
evidence.

\textbf{Input:}
\begin{itemize}[nosep,leftmargin=*]
  \item Tool description: \{tool\_description\}
  \item Usage evidence from \{n\_queries\} tasks:
    \{evidence\_text\}
  \item Existing affordance entries to avoid
    duplicates: \{existing\_affordance\}
  \item Existing boundary entries to avoid
    duplicates: \{existing\_boundaries\}
  \item Existing co-usage entries to avoid
    duplicates: \{existing\_co\_usage\}
\end{itemize}
\textbf{Abstraction rule:} Produce
environment-invariant entries grounded in usage
evidence. Do NOT reference specific filenames,
page IDs, table names, issue numbers, schema
fields, literal values, paths, or task identifiers
inside the entry text.
\textbf{Output (JSON):}
\begin{itemize}[nosep,leftmargin=*]
  \item \texttt{new\_affordance}: valid uses that
    $t_i$ reliably supports.
  \item \texttt{new\_boundaries}: regimes where
    $t_i$ becomes unreliable or infeasible.
  \item \texttt{new\_co\_usage}: recurrent
    compositions with peer tools.
\end{itemize}
\noindent\texttt{\{} \\
\quad \texttt{"new\_affordance": [\{"text": "<affordance>",} \\
\quad\quad \texttt{"source\_queries": ["q1","q2"]\}],} \\
\quad \texttt{"new\_boundaries": [\{"text": "<boundary>",} \\
\quad\quad \texttt{"source\_queries": ["q1"]\}],} \\
\quad \texttt{"new\_co\_usage": [\{"text": "<co-usage pattern>",} \\
\quad\quad \texttt{"related\_tools": ["tool\_a","tool\_b"],} \\
\quad\quad \texttt{"source\_queries": ["q1","q2"]\}]} \\
\texttt{\}}
Only include entries backed by usage evidence.
Every entry must cite \texttt{source\_queries}.
If there is nothing new to add, return empty lists.
\end{tcolorbox}

\begin{tcolorbox}[
  colback=lightgray!20,
  colframe=darkgray!80,
  title=Tool-Strategy Graph Induction Prompt,
  breakable,
  left=2mm, right=2mm,
  top=1mm, bottom=1mm]
\label{sec:prompt:tool-strategy-induction}
You are analyzing task-level tips from multiple
tasks to find recurring planning patterns.
\textbf{Goal:} Update the Tool-Strategy Graph
$\mathcal{G}_{\text{strat}}$ by distilling
recurring task-level tips into strategy nodes
that capture high-level orchestration principles.
\textbf{Input:}
\begin{itemize}[nosep,leftmargin=*]
  \item Task-level tips collected from
    \{n\_queries\} tasks: \{all\_tips\_text\}
  \item Existing strategy nodes to avoid
    duplicates: \{existing\_strategies\}
  \item Minimum recurrence threshold:
    \{threshold\} source queries
\end{itemize}
\textbf{Selection rules:}
\begin{itemize}[nosep,leftmargin=*]
  \item Keep only planning patterns that recur
    across multiple tasks; discard single-tool
    properties.
  \item Prefer high-level orchestration principles
    over execution-level tool tips.
  \item Each strategy node must be backed by at
    least \{threshold\} source queries.
  \item Do not repeat existing strategy nodes.
\end{itemize}
\textbf{Output (JSON):}
\begin{itemize}[nosep,leftmargin=*]
  \item \texttt{new\_strategy}: recurring
    high-level orchestration principles with
    supporting query IDs.
\end{itemize}
\noindent\texttt{\{} \\
\quad \texttt{"new\_strategy": [\{"text": "<orchestration principle>",} \\
\quad\quad \texttt{"source\_queries": ["q1","q2","q3"]\}]} \\
\texttt{\}}
Only include genuinely new strategy nodes. Return
an empty list if nothing qualifies.
\end{tcolorbox}

\begin{tcolorbox}[
  colback=lightgray!20,
  colframe=darkgray!80,
  title=Dynamic Memory Traversal Prompt,
  breakable,
  left=2mm, right=2mm,
  top=1mm, bottom=1mm]
\label{sec:prompt:dynamic-memory-traversal}
You are the navigator in Dynamic Memory Traversal.
You have access to a tool memory graph
$\mathcal{M}_\mathcal{T} = (\mathcal{G}_{\text{trace}},
\mathcal{G}_{\text{cap}}, \mathcal{G}_{\text{strat}})$
induced from past task rollouts.
\textbf{Goal:} Given a NEW TASK $Q$, perform a
graph walk over $\mathcal{M}_\mathcal{T}$ and
compress the gathered context into
task-conditioned guidance $g(Q)$ for the user
agent.
\textbf{Traversal operations:}
\begin{itemize}[nosep,leftmargin=*]
  \item \texttt{ReadTrace(q)}: read the task,
    agent-neutral trace, and task-level tips of
    trace node $q$.
  \item \texttt{Expand(q)}: follow trace-graph
    edges to similar trace nodes.
  \item \texttt{ReadTool(t)}: read affordance,
    boundary, and co-usage entries from tool
    node $t$.
  \item \texttt{ReadStrategy(q)}: read strategy
    nodes supported by trace node $q$.
  \item \texttt{Done}: stop traversal and emit
    the final task-conditioned guidance $g(Q)$.
\end{itemize}
\textbf{Traversal policy:}
\begin{itemize}[nosep,leftmargin=*]
  \item Start from $k_r$ nearest Tool-Trace
    nodes as the initial context $C_0$.
  \item Use graph neighbors before global
    search.
  \item Operate under a fixed read budget; call
    \texttt{Done} when the context is sufficient
    or the read budget is exhausted.
  \item Each operation call MUST include a
    \texttt{thought} field explaining the choice.
\end{itemize}
\textbf{Output (JSON via \texttt{Done}):} the
emitted guidance $g(Q)$ has four sections, each
grounded in the corresponding layer of
$\mathcal{M}_\mathcal{T}$.
\begin{itemize}[nosep,leftmargin=*]
  \item \texttt{seed\_candidates}: the initial
    Tool-Trace seeds, each with its qid,
    embedding similarity to $Q$, and a one-line
    summary.
  \item \texttt{playbook}: an ordered sequence
    of \texttt{(tool, rationale)} pairs
    synthesized from retrieved agent-neutral
    traces and adapted to $Q$. Each
    \texttt{tool} must be a real MCP tool name.
  \item \texttt{strategy}: applicable
    orchestration principles drawn from strategy
    nodes in $\mathcal{G}_{\text{strat}}$, each
    with its source strategy-node id.
  \item \texttt{tool\_tips}: per-tool hints
    relevant to $Q$, drawn from affordance,
    boundary, and co-usage entries; include
    negative tips (\emph{when not to use}) when
    boundary entries apply.
\end{itemize}
\noindent\texttt{\{} \\
\quad \texttt{"seed\_candidates": [\{"qid": "<qid>",} \\
\quad\quad \texttt{"sim": <float>,} \\
\quad\quad \texttt{"summary": "<one-line summary>"\}],} \\
\quad \texttt{"playbook": [\{"tool": "<real\_mcp\_tool>",} \\
\quad\quad \texttt{"rationale": "<role of this step for $Q$>"\}],} \\
\quad \texttt{"strategy": [\{"text": "<orchestration principle>",} \\
\quad\quad \texttt{"source\_strategy\_id": "<strategy\_node\_id>"\}],} \\
\quad \texttt{"tool\_tips": [\{"tool": "<real\_mcp\_tool>",} \\
\quad\quad \texttt{"tip": "<task-relevant hint>"\}]} \\
\texttt{\}}
Omit generic advice. Never recommend
unavailable tools. If no past task is relevant,
return empty lists for all four sections.
\end{tcolorbox}





\end{document}